\documentclass{ieeeaccess}
\usepackage{cite}
\usepackage{amsmath,amssymb,amsfonts}
\usepackage{algorithmic}
\usepackage{graphicx,url}
\usepackage{textcomp}
\usepackage{multirow}
\usepackage{bbm,color}
\usepackage[caption=false, font=footnotesize]{subfig}
\usepackage[ruled,vlined]{algorithm2e}
\def\BibTeX{{\rm B\kern-.05em{\sc i\kern-.025em b}\kern-.08em
    T\kern-.1667em\lower.7ex\hbox{E}\kern-.125emX}}
\begin{document}
\history{Date of publication xxxx 00, 0000, date of current version xxxx 00, 0000.}
\doi{10.1109/ACCESS.2017.DOI}

\title{UAV-based Crowd Surveillance in Post COVID-19 Era}
\author{\uppercase{Nizar~Masmoudi}\authorrefmark{1},~\uppercase{Wael~Jaafar}\authorrefmark{2},~\IEEEmembership{Senior Member, IEEE},~\uppercase{Safa Cherif}\authorrefmark{3},~\uppercase{Jihene~Ben~Abderrazak}\authorrefmark{4} and \uppercase{Halim~Yanikomeroglu}\authorrefmark{5}, \IEEEmembership{Fellow, IEEE}}
\address[1]{ESPRIT School of Engineering, Tunis, Tunisia (e-mail: nizar.masmoudi@esprit.tn)}
\address[2]{Department of Systems and Computer Engineering, Carleton University, Ottawa, Canada (e-mail: waeljaafar@sce.carleton.ca)}
\address[3]{ESPRIT School of Engineering, Tunis, Tunisia (e-mail: safa.zhiouacherif@esprit.tn)}
\address[4]{ESPRIT School of Engineering, Tunis, Tunisia (e-mail: jihene.benabderrazek@esprit.tn)}
\address[5]{Department of Systems and Computer Engineering, Carleton University, Ottawa, Canada (e-mail: halim@sce.carleton.ca)}
\tfootnote{This work is funded in part by Huawei Canada and in part by the Natural Sciences and Engineering Research Council of Canada (NSERC).\\
\textcolor{red}{This work is accepted for publication in IEEE Access. Copyright will be transferred to IEEE at the final publishing stage.  
}}

\markboth
{Author \headeretal: Preparation of Papers for IEEE TRANSACTIONS and JOURNALS}
{Author \headeretal: Preparation of Papers for IEEE TRANSACTIONS and JOURNALS}

\corresp{Corresponding author: Wael Jaafar (e-mail: waeljaafar@sce.carleton.ca).}

\begin{abstract}
    To cope with the current pandemic situation and reinstate pseudo-normal daily life, several measures have been deployed and maintained, such as mask wearing, social distancing, hands sanitizing, etc. Since outdoor cultural events, concerts, and picnics, are gradually allowed, a close monitoring of the crowd activity is needed to avoid undesired contact and disease transmission. In this context, intelligent unmanned aerial vehicles (UAVs) can be occasionally deployed to ensure the surveillance of these activities, that health restriction measures are applied, and to trigger alerts when the latter are not respected. Consequently, we propose in this paper a complete UAV framework for intelligent monitoring of post COVID-19 outdoor activities. Specifically, we propose a three steps approach. In the first step, captured images by a UAV are analyzed using machine learning to detect and locate individuals. The second step consists of a novel coordinates mapping approach to evaluate distances among individuals, then cluster them, while the third step provides an energy-efficient and/or reliable UAV trajectory to inspect clusters for restrictions violation such as mask wearing. Obtained results provide the following insights: 1) Efficient detection of individuals depends on the angle from which the image was captured, 2) coordinates mapping is very sensitive to the estimation error in individuals' bounding boxes, and 3) UAV trajectory design algorithm 2-Opt is recommended for practical real-time deployments due to its low-complexity and near-optimal performance.
\end{abstract}


\titlepgskip=-15pt

\maketitle

\section{Introduction} \label{sec:introduction}
    \PARstart{S}{ince} early 2020, the world has been living a multi-wave COVID-19 pandemic, which forced populations into quarantine and limited movement episodes. However, such a situation cannot last forever for the sake of mental health. Also, with the progress of vaccination, health restrictions are being loosened precociously.
    Indeed, the World Health Organization started a huge campaign to spread awareness about COVID-19 and share health measures to minimize infections. Besides the wearing of masks and constant use of sanitizers to reduce viral transmission, social distancing rules were enforced. Specifically, individuals should always keep a safety distance of more than 1 or 2 meters (depending on the country) away from each other. This sparked a new interest in crowd surveillance not only to overcome the current situation but also to prevent future pandemics.
    
    Several intelligent solutions were proposed to evaluate social distancing between people. For instance, Yang \textit{et al.} described in \cite{yang2020visionbased} a system that sends a multimedia message to alert the crowd when social distancing is disrespected. Fed with a surveillance camera video-stream, their solution localizes pedestrians using a pre-trained conventional object detector. Then, inverse homography transformation is used to map image pedestrians' coordinates with real-world locations and to evaluate distances among them.   
    Punn \textit{et al.} adopted in \cite{punn2021monitoring} a similar setting, where they trained a YOLOv3 model alongside DeepSORT to localize detected people in a surveillance camera video-stream. Then, real-world coordinates were estimated using bounding boxes around detected individuals. 
    
    In addition to works relying on captured images/videos by fixed surveillance cameras, unmanned aerial vehicle (UAV) based surveillance systems have been also considered. Indeed, a large number of organizations and industries are leveraging the UAV technology to automate complicated tasks. Specifically, UAVs are capable of flying at high altitudes to expand their range of view and take advantage of obstacle-free spaces. For instance, modern agriculture uses UAVs for soil health scans, fertilization, and crop monitoring. Also, law enforcement organizations use UAVs for crime investigation and crowd control, especially during large public events. In the context of COVID-19,  
    Somaldo \textit{et al.} leveraged UAVs for social distance monitoring \cite{9357040}. Using the UAV's ventral camera, the system feeds aerial images into a trained YOLOv3-based object detector. Based on the area expansion principle, the authors developed a coordinates mapping algorithm and evaluated social distancing conditions between detected pedestrians.
    Although Yang \textit{et al.} managed in \cite{yang2020visionbased} to achieve accurate results, their system is sensitive to the environment setting. Furthermore, the use of fixed cameras to monitor wide open areas \cite{yang2020visionbased,punn2021monitoring} is not efficient due to the small covered area by a single camera. This issue is bypassed in \cite{9357040} through the use of flexibly flying UAVs. Nevertheless, the UAV's camera has generally a limited vision range, thus requiring either flying over long distances to monitor a wide area, which consumes a significant amount of energy, or deploying several UAVs simultaneously.
    
    Motivated by the aforementioned issues, we propose in this paper a complete UAV-based framework for outdoor crowd surveillance in the post COVID-19 era. The latter follows three steps as follows: 1) Given a single UAV that hovers and captures images on a crowded open area, the images stream is run through an object detector to fit bounding boxes around people. Due to the low-efficiency of detection model with heterogeneous individuals, i.e., standing or seated adults, kids, etc., we propose a bounding box correction process.  
    2) Corrected bounding boxes are then used for coordinates mapping into a 1:1 scale coordinates system. Subsequently, individuals are clustered with respect to the social distancing conditions. 3) Each cluster is attributed an infection risk score, calculated based on distances between individuals within the same cluster. Then, the UAV is deployed closer to clusters in order to check other conditions, such as mask wearing, identity, or vaccination pass. Several trajectory design algorithms are evaluated, with the objective trading-off between prioritizing clusters with high risks and reducing energy consumption. 
    To the best of our knowledge, this is the first work that provides a complete UAV-based surveillance framework for post COVID-19 circumstances.
    The contributions of the paper are given as follows:
    \begin{enumerate}
    	\item We present a complete UAV based surveillance framework that combines three aspects, namely individuals detection and localization, coordinates mapping, distance evaluation, individuals clustering, and UAV trajectory design.
    	\item Differently from previous works, we propose an adaptation of the individuals localization algorithm to support more efficiently heterogeneous cases, including standing/seated adults and kids. Specifically, since low altitude wide-angle captured images raise bounding box errors with seated adults and kids detection, we propose a bounding box correction technique to reduce its effect on coordinates mapping.
    	\item Unlike most state-of-the-art works that rely on additional information, e.g., camera characteristics, capturing angle, GPS information, landmarks, etc., in order to map image coordinates to real-world ones, our work proposes a coordinates mapping solution based solely on the set of images, without involving any other information.
    	
    	\item Finally, we propose an efficient UAV trajectory design to further investigate the crowd, for instance, to verify mask wearing restrictions. The UAV path is composed of two components: The first defines the order of visit for each individuals cluster, while the second determines how the individuals within a cluster are inspected. 
    \end{enumerate}
    
    The remaining of the paper is organized as follows. Section \ref{sec:rel-work} reviews the related works. Section \ref{sec:human-det} describes the people detection phase. Section \ref{sec:img-analysis-cls} details how to map detected individuals to real-world locations, and the clustering approach. Section \ref{sec:uav-traj-opt} formulates and solves the UAV trajectory planning problem. 
    Then, Section \ref{sec:res-dis} presents and discusses the results. Finally, Section \ref{sec:conc-future-wrks} concludes the paper.
    
\section{Related Work} \label{sec:rel-work}
    We present in this section the most relevant works to our framework's steps, namely human detection, social distance monitoring, and UAV trajectory optimization. A summary is presented in Table \ref{tab:rel-table}.
    
    \subsection{Human detection} \label{sec:obj-det}
        Human detection algorithms are mainly classified into two categories: Two-stage and one-stage object detectors.
            
        Two-stage object detectors split the detection process into two steps. The first step is a convolutional region network, a.k.a, region based convolutional neural networks (R-CNN). It extracts feature maps from the input image and provides a certain number of regions called regions of interest (ROI). ROIs are then classified in the second step via a dense neural network to provide the final bounding boxes that locates individuals. The R-CNN is the foundation of most two-stage object detectors \cite{girshick2014rich}. Faster R-CNN, proposed in \cite{ren2016faster}, managed to reach a mean average precision (mAP) of 69\% on the PASCAL visual object classes (VOC) 2007 dataset in \cite{article}.
            
        In contrast, one-stage object detectors have a fully convolutional architecture. The latter behaves as a single regression model predicting bounding boxes and class probabilities from the entire image, which provides training and inference speed, and the capability to generalize through different scenarios. Redmon \textit{et al.} were the first to introduce this architecture in YOLOv1 \cite{redmon2016look}. Although YOLOv1 achieved only 63\% mAP on PASCAL VOC 2007 dataset, its inference speed was significantly higher than that of two-stage object detectors, while the recently proposed YOLOv4 in \cite{bochkovskiy2020yolov4} achieved a higher mAP with respect to the fully convolutional architecture.
            
        Human detection can be seen as part of a bigger object detection problem. For instance, authors of \cite{punn2021monitoring} suggested to combine a YOLOv3-based model with DeepSORT. Their algorithm managed to reach a mAP of 84.6\% when trained and evaluated on the Open Image Dataset. 
        Several other works focused on object detection in aerial images \cite{pailla2019object}\nocite{tang2020penet,albaba2020synet}--\cite{yu2020resolving}. They relied on the VisDrone2019 dataset for training and testing, but using different machine learning models. For instance, authors of
        \cite{pailla2019object} trained a CenterNet algorithm with a large input image resolution (2048 x 2048) in order to avoid the loss of discriminatory features for small objects. Their model reached mAP of 65\% on the validation subset on human classes, i.e., \textit{people} and \textit{pedestrian}. Also, the authors of \cite{tang2020penet} proposed a novel network structure called PENet (Points Estimated Network) that achieved 41\% mAP on the validation subset on human classes, while 
        authors in \cite{albaba2020synet} combined Cascaded R-CNN with CenterNet through weighted box ensemble to design a new network called SyNet. Their proposed model detected humans with mAP of 43\%. In contrast, Yu \textit{et al.} proposed in \cite{yu2020resolving} an algorithm based on Faster R-CNN, called DSHNet. Between \textit{people} and \textit{pedestrian} classes, their model recorded a mAP of only 19.5\%. 
        The aforementioned works demonstrate the difficulty to efficiently detect humans through aerial imagery. Indeed, 
        VisDrone2019, amongst only few other UAV-based image datasets, offers different perspectives, scales, and angles of capture, which complicates the detection algorithm generalization.
        
    \subsection{Social distance monitoring}
        Social distance monitoring is a complex task that relies essentially on mapping image coordinates to real-world coordinates. In this context, several works investigated mapping functions. In \cite{yang2020visionbased}, authors used the inverse homography transformation matrix, while \cite{punn2021monitoring} relied on the convex lens geometric properties and the dimensions of bounding boxes to estimate the depth of each detected individual. Moreover, authors of \cite{10.1007/978-3-642-40567-9_1} focused on implementing a camera calibration method to estimate intrinsic and extrinsic camera parameters. They developed a corner detection algorithm to establish the real-world coordinates system. 
        In \cite{9357040}, individuals' real coordinates were estimated using the characteristics of the used UAV-mounted camera and a calibration constant defined based on the area expansion principle, while homography based estimation was realized in \cite{BABINEC2016152} by relaying on capturing at least four landmarks.
    
        Although the aforementioned works successfully mapped image coordinates to real-world ones, most of them were built under several assumptions and camera parameters knowledge. For instance, \cite{yang2020visionbased, 10.1007/978-3-642-40567-9_1} and \cite{BABINEC2016152} require the presence of essential landmarks in order to estimate their mapping function. These assumptions cannot be supported with others datasets, such as VisDrone2019.
        
    \subsection{UAV trajectory optimization}
        With the growing interest in leveraging UAVs for automating complex tasks, 
        UAV trajectory optimization is one of the most critical issues. In \cite{zhang_celik_dang_shihada_2021}, the authors proposed UAV path design aiming to minimize energy consumption, data transmission, and coverage fairness, while 
        authors of \cite{coupechoux2021optimal} solved optimally the UAV trajectory problem, targeting minimum cellular data offloading costs, using Hamiltonian-Jacobi equations (H\&J eq.). In
        \cite{fountoulakis2020uav}, the authors investigated the UAV path planning problem to maximize the 
        total score collected from specific regions within a mission time constraint.
        Their solution inspired  from the nearest neighbor (NN) algorithm built a trajectory by progressively adding nodes that maximize the ration reward by distance.
        Finally, authors of \cite{Ghdiri2021} proposed several algorithms in the offline and online settings to maximize UAV data collection from ground sensors and within time deadlines. Their offline solutions included Tabu search, simulated annealing (SA), and guided local search (GLS), while the online solutions involved reinforcement learning (RL) algorithms.
        
    The following Sections III, IV, and V, expose our crowd surveillance framework, as presented in Fig. \ref{fig:summary}.
    For the sake of clarity, we summarize in Table \ref{tab:symbols} the symbols used in the remaining of the paper, along their descriptions.
    
    \begin{table*}
    \centering
    \caption{Summary of related works.}
        \label{tab:rel-table}
        \begin{tabular}{|l|c|c|c|c|c|c|c|} 
            \hline
            \multirow{2}{*}{Paper} & \multicolumn{3}{c|}{Focus} & \multicolumn{4}{c|}{Features} \\ 
            \cline{2-8} & \begin{tabular}[c]{@{}c@{}}Human\\detection\end{tabular} & \begin{tabular}[c]{@{}c@{}}Coordinates \\mapping\end{tabular} & \begin{tabular}[c]{@{}c@{}}UAV trajectory \\optimization\end{tabular} & \multicolumn{1}{c|}{UAV-based} & \multicolumn{1}{c|}{Detection model} & \begin{tabular}[c]{@{}c@{}}Coords. mapping\\ method\end{tabular} & \begin{tabular}[c]{@{}c@{}}UAV trajectory \\algorithm\end{tabular}  \\ 
            \hline
            Punn \textit{et al.} \cite{punn2021monitoring} & \checkmark & \checkmark & -- & -- & YOLOv3 \& Deepsort & \begin{tabular}[c]{@{}c@{}}Bounding box\\ based\end{tabular} & -- \\ 
            \hline
            Pailla~\textit{et al.} \cite{pailla2019object} & \checkmark & -- & -- & \checkmark & CenterNet & -- & -- \\ 
            \hline
            Tang~\textit{et al.} \cite{tang2020penet} & \checkmark & -- & -- & \checkmark & PENet & -- & -- \\ 
            \hline
            Albaba~\textit{et al.} \cite{albaba2020synet} & \checkmark & -- & -- & \checkmark & SyNet & -- & -- \\ 
            \hline
            Yu~\textit{et al.} \cite{yu2020resolving} & \checkmark & -- & -- & \checkmark & DSHNet & -- & -- \\ 
            \hline
            Yang~\textit{et al.} \cite{yang2020visionbased} & -- & \checkmark & -- & -- & -- & \begin{tabular}[c]{@{}c@{}}Inverse homography\\transformation\end{tabular} & -- \\ 
            \hline
            Somaldo~\textit{et al.} \cite{9357040} & -- & \checkmark & -- & \checkmark & -- & -- & -- \\ 
            \hline
            Babinec~\textit{et al.} \cite{BABINEC2016152} & -- & \checkmark & -- & \checkmark & -- & \begin{tabular}[c]{@{}c@{}}Inverse homography\\transformation\end{tabular} & -- \\ 
            \hline
            Siswantoro~\textit{et al.} \cite{10.1007/978-3-642-40567-9_1} & -- & \checkmark & -- & -- & -- & Camera calibration & -- \\ 
            \hline
            Zhang~\textit{et al.} \cite{zhang_celik_dang_shihada_2021} & -- & -- & \checkmark & \checkmark & -- & -- & RL \\ 
            \hline
            Coupechoux~\textit{et al.} \cite{coupechoux2021optimal} & -- & -- & \checkmark & \checkmark & -- & -- & H\&J eq. \\ 
            \hline
            %Chen~\textit{et al.} \cite{CHEN2016184} & -- & -- & \checkmark & \checkmark & -- & -- & CFO-GA \\ 
            %\hline
            Fountoulakis~\textit{et al.} \cite{fountoulakis2020uav} & -- & -- & \checkmark & \checkmark & -- & -- & NN-based \\ 
            \hline
            Ghdiri~\textit{et al.} \cite{Ghdiri2021} & -- & -- & \checkmark & \checkmark & -- & -- & Tabu,SA,GLS,RL\\ 
            \hline
            This work & \checkmark & \checkmark & \checkmark & \checkmark & Scaled YOLOv4 & 
            \begin{tabular}[c]{@{}c@{}}Bounding box\\ based\end{tabular}
             & 2-Opt, GA, ACO \\
             \hline
        \end{tabular}
    \end{table*}
    
    \Figure[t!](topskip=0pt, botskip=0pt, midskip=0pt)[width=0.99\linewidth]{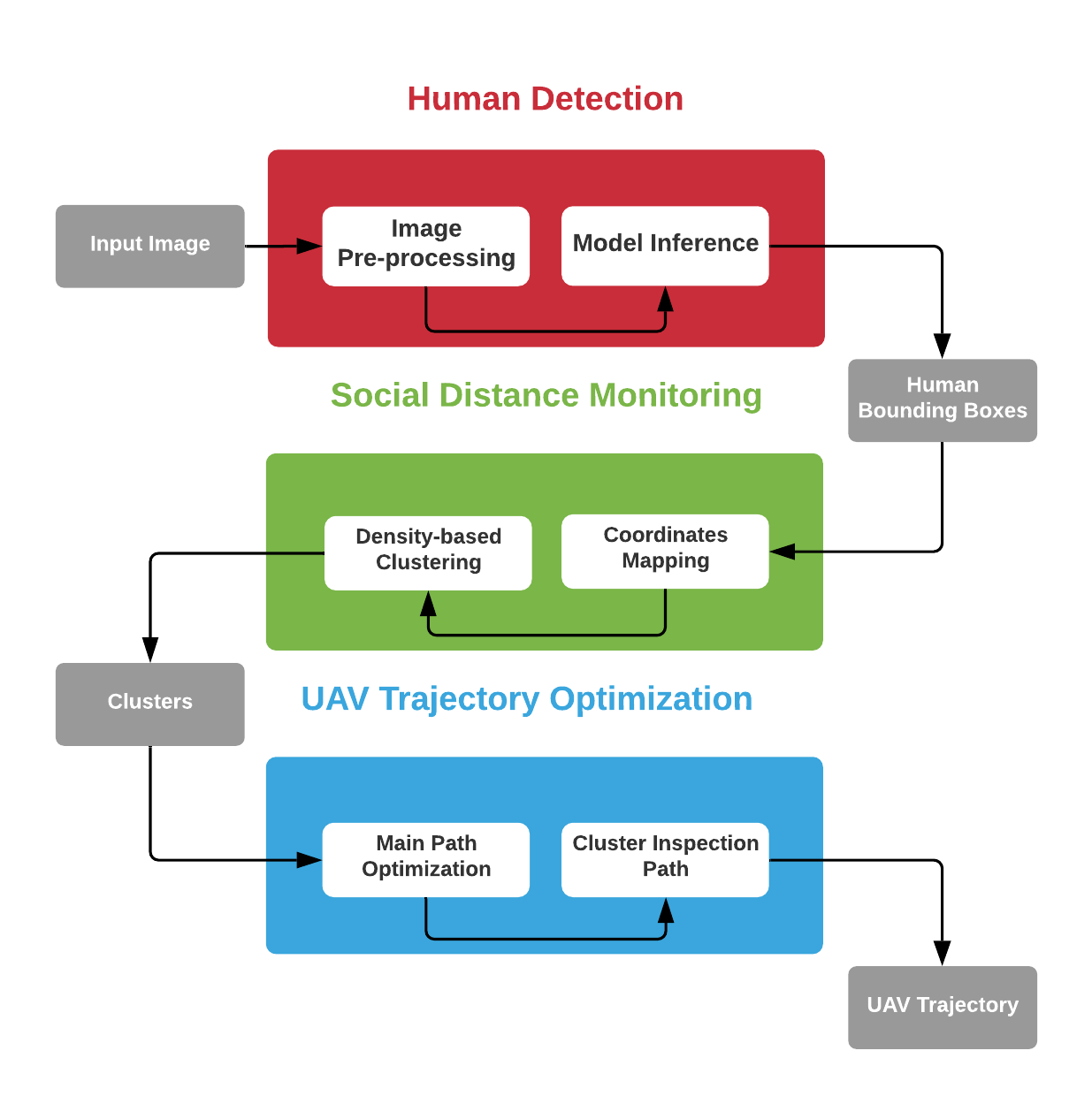}{Proposed UAV framework for intelligent monitoring of post COVID-19 activities.\label{fig:summary}}
    
    \begin{table}[t]
        \centering
        \caption{List of symbols.}
        \label{tab:symbols}
        \begin{tabular}{ll}
            \textbf{Symbol} & \textbf{Description}  \\
            \hline
            $(x, y)$ & Image plane coordinates system (in pixels) \\
            $(\bar{x}, \bar{y})$ & Image plane coordinates system (in meters)\\
            $(\hat{x}, \hat{y}, \hat{z})$ & Real-world coordinates system (in meters) \\
            $h$ & Bounding box height (in pixels) \\
            $w$ & Bounding box width (in pixels) \\
            $h_c$ & Corrected bounding box height (in pixels) \\
            $\bar{h}$ & Corrected bounding box height (in meters) \\
            $f$ & Focal length of camera lens \\
            $\hat{h}$ & Assumed height of a perfectly standing adult \\
            $\mathcal{G}$ & Complete directed graph of clusters \\
            $\mathcal{C}$ & Set of clusters \\
            $\mathcal{E}$ & Set of edges connecting clusters of $\mathcal{C}$ \\
            $\mathbf{F}$ & Cost matrix of edges in $\mathcal{E}$ \\
            $f_{ij}$ & Total cost of edge $(i, j) \in \mathcal{E}$\\
            $f^p_{ij}$ & Priority cost of edge $(i, j) \in \mathcal{E}$ \\
            $f^e_{ij}$ & Energy consumption cost of edge $(i, j) \in \mathcal{E}$ \\
            $\alpha$ & Weight of cost $f_{ij}^p$ \\
            $\mathbf{c}_i$ & Coordinates of cluster $i$ in system $(\hat{x},\hat{y})$ \\
            $\lambda_i$ & Infection risk score of cluster $i$ \\
            $N_i$ & Number of individuals in cluster $i$ \\
            $t^*$ & Optimized UAV trajectory \\
            $\mathcal{W}$ & Set of convex hull points related to a cluster \\
            $d_S$ & Safety distance \\
        \end{tabular}
    \end{table}

\section{Human Detection} \label{sec:human-det}
    In this section, we present and explain the setup for human detection.
    
    \subsection{Aerial images dataset}
        Amongst only few public datasets captured by camera-equipped UAVs, VisDrone2019 \cite{zhuvisdrone2018} stands out as one of the largest, most diverse and carefully annotated benchmark datasets. VisDrone2019 was provided by AISKEYE team at the Lab of Machine Learning and Data Mining in Tianjin University, China and was designed for multiple tasks with over 10 different object classes (people, pedestrian, car, van, truck, \dots). With over 8,000 images, VisDrone2019 images were captured from different altitudes and angles and under diverse circumstances, namely night, rain, lens flare, and motion blur. Besides, the captured images vary in size from 480 x 360 up to 2000 x 1500.
        
        VisDrone2019 was designed for machine learning, hence the images were carefully split into a training set (6,471 images), a validation set (1,610 images) and a test set (548 images).
        
\begin{figure*}[ht]
            \centering
          \subfloat[Image 0000011\_00234\_d\_0000001.]{\includegraphics[width=0.45\linewidth]{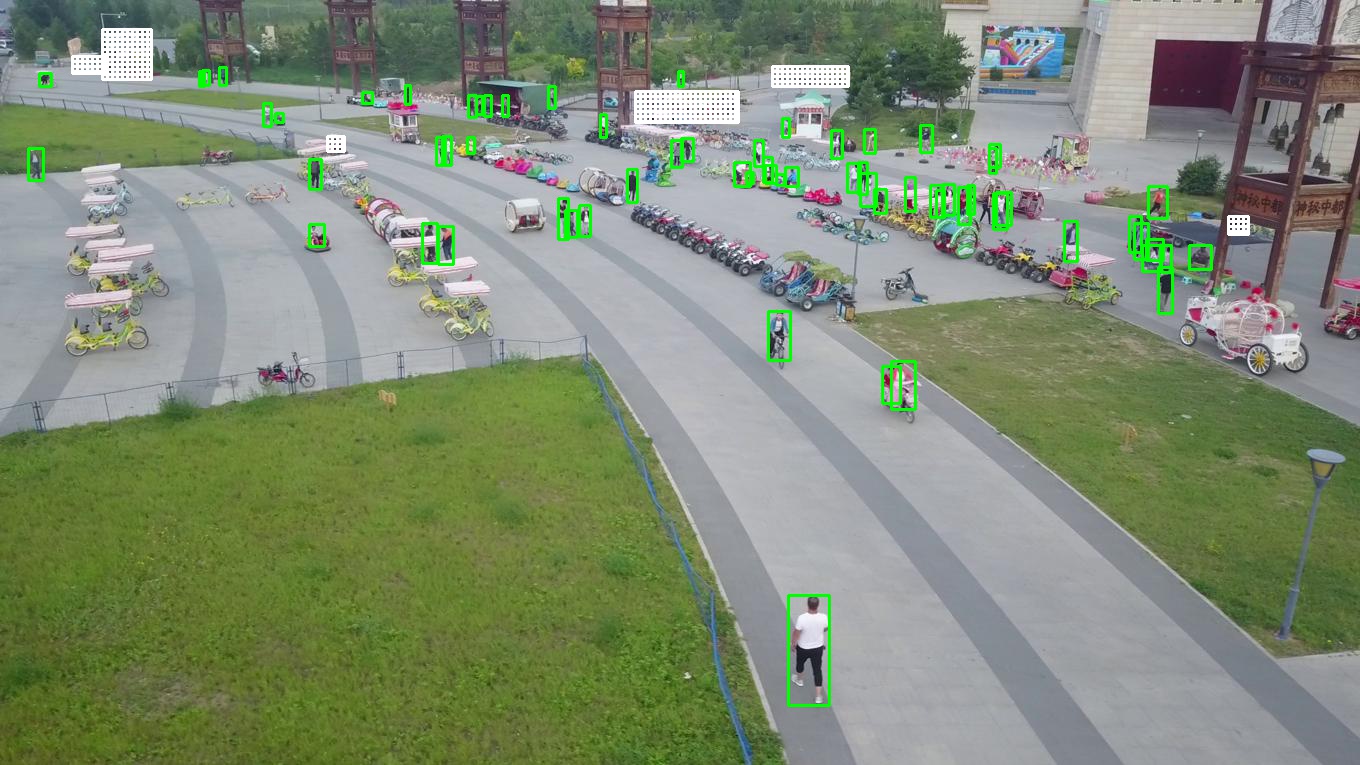}}%
            \hfil
            \subfloat[Image 0000031\_02000\_d\_0000041.]{\includegraphics[width=0.45\linewidth]{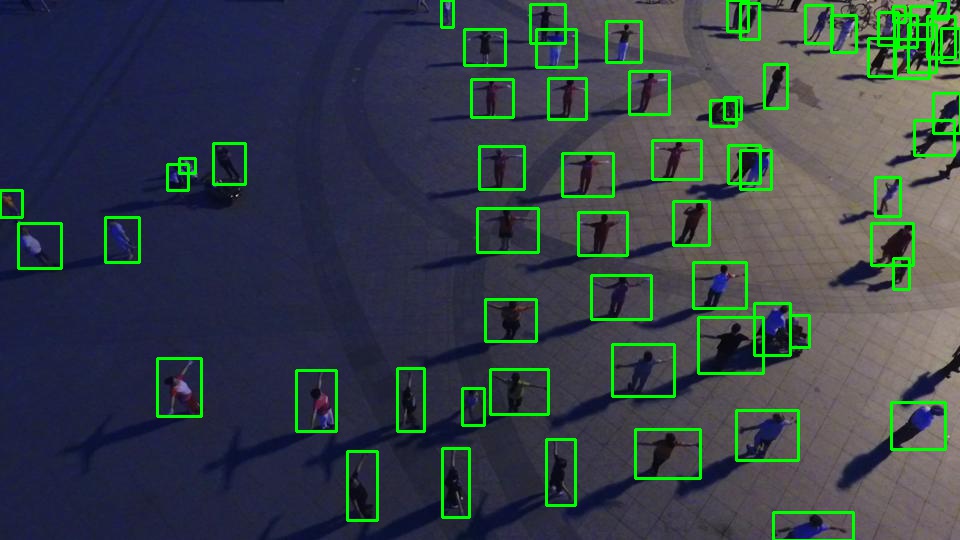}}%
            \vskip\baselineskip
            \subfloat[Image 0000129\_02411\_d\_0000138.]{\includegraphics[width=0.45\linewidth]{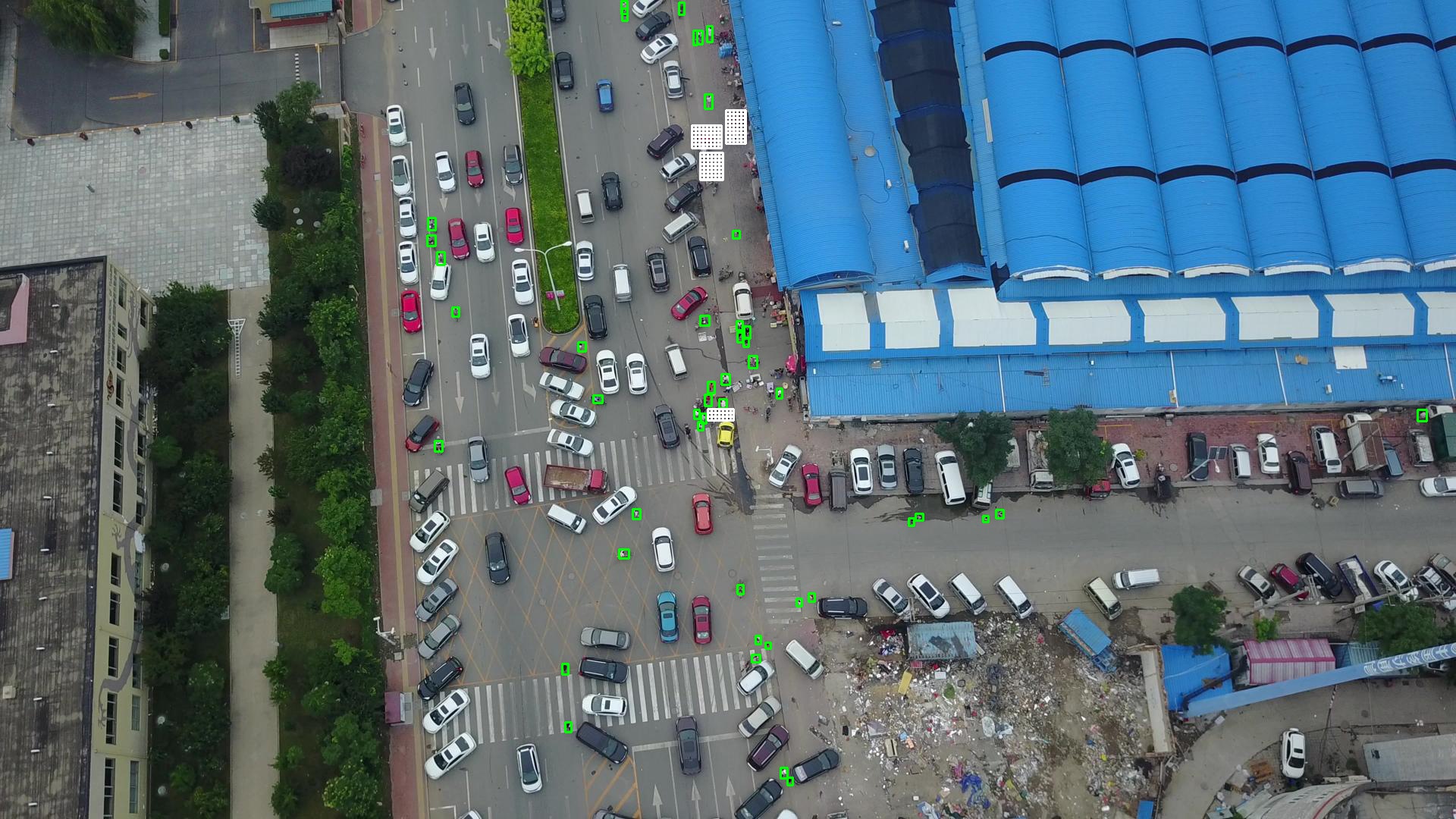}}%
            \hfil
            \subfloat[Image 0000074\_08777\_d\_0000017.]{\includegraphics[width=0.45\linewidth]{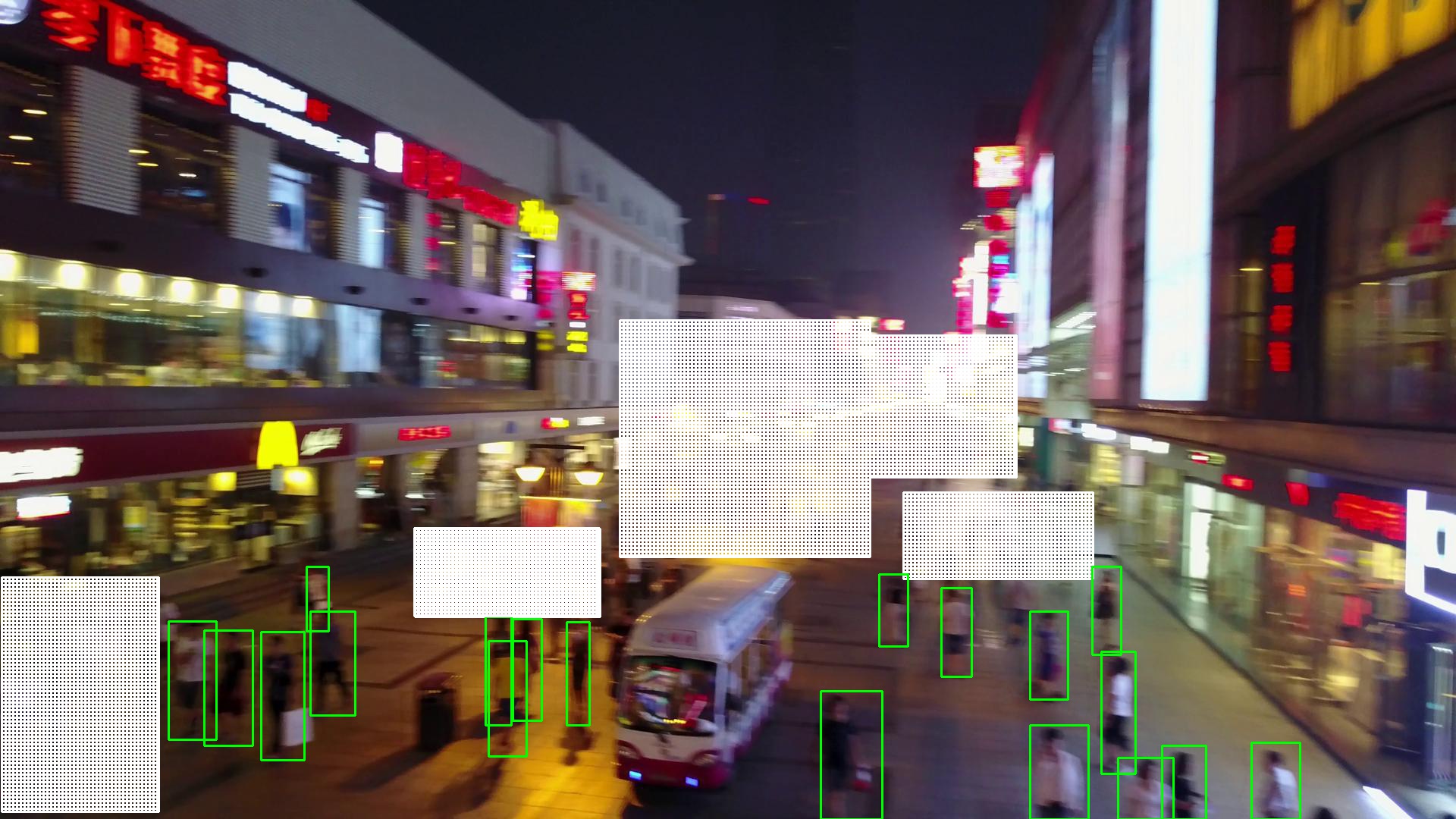}}%
            
            \caption{Annotated samples from VisDrone2019 dataset (Green rectangles identify humans; White grids are ignored regions).}
            \label{fig:visdrone-samples}
        \end{figure*}

        AISKEYE team provided comprehensive annotations for each image presenting useful information, namely the bounding boxes' locations as well as their dimensions, the encoded object category and the truncation and occlusion ratios of each object. Given the annotation complexity of some regions due to a low resolution and/or a dense crowded area, AISKEYE team identified portions of images to be ignored while training or evaluation in order to avoid misleading the model. Annotated samples from the VisDrone2019 dataset are presented in Fig. \ref{fig:visdrone-samples}.
        
        % \Figure[t!](topskip=0pt, botskip=0pt, midskip=0pt)[width=\linewidth]{Figs/Dataset-samples}{Annotated samples from VisDrone2019 dataset (Green rectangles identify humans; White grids are ignored regions).\label{fig:visdrone-samples}}
        
    \subsection{Pre-processing} \label{sec:data-pre-proc}
        First of all, ignored regions that were likely to disturb the model's development were replaced with white noise as new pixels were drawn from a Gaussian distribution.
        
        As mentioned previously, VisDrone2019 images can be very large (up to 2000 x 1500 pixels). In order to efficiently extract features from small objects (5 x 15 pixels), it is imperative that the network's input dimensions at least match the dimensions of these large images in order to avoid input compression and the loss of features crucial for detection. Evidently, large networks require more computational power. Due to the inaccessibility to high-end hardware, we were obligated to split the images into 4 portions each lowering, therefore, the required size of the network. Each portion was, then, resized to 736 x 736 pixels.
        
        The splitting process generates a heavy amount of negative samples. Although these images may present an addition to the training set, their influence on the model's precision is very minimal compared to the amount of training time they cost. In fact, object detectors consider each group of pixels that does not belong to the annotated objects as negatives.
        
        Given that the objective of this task is to only detect humans, \textit{pedestrian} and \textit{people} classes were merged into one class while other objects were omitted from the dataset.
        
    \subsection{Scaled YOLOv4}
        For human detection, we opted in this work for the recent Scaled YOLOv4 model \cite{wang2021scaledyolov4}. Authors of \cite{wang2021scaledyolov4} proposed initially the YOLOv4 model, which inherits the fully convolutional architecture of its ancestors. Although this object detection model achieved state-of-the-art results, Bochkovskiy \textit{et al.} further enhanced their architecture by readjusting the backbone, adding more Cross Stage Partial (CSP) connections and scaling up the model to get the best speed/accuracy trade-off. This novel architecture, called Scaled YOLOv4, has been proven to achieve high detection results. Indeed, as shown in \cite{wang2021scaledyolov4}, ground-breaking results of 55.4\% in average precision (AP) on the Microsoft Common Objects in Context (MS COCO) dataset with a relatively high inference speed is realized.
        
    \subsection{Model training}
        In order to train the model for our dataset, an implementation of scaled YOLOv4 on GitHub is cloned to a Google Colaboratory environment \cite{darknet,colab}. The GPU provided by the environment is NVIDIA Tesla T4. 
        
        For our work, we initialize scaled YOLOv4 with the pre-trained weights on the MS COCO dataset. The learning rate starts with a very low value and gradually increases during a short warm up phase until it reaches 0.001. The image input size is set to $736 \times 736$ pixels, and a preset of 9 bounding boxes were calculated with the K-means algorithm on the training subset \cite{1017616}. The algorithm uses the complete intersection over union (CIoU) loss \cite{zheng2019distanceiou} and Nesterov accelerated gradient for back-propagation \cite{botev2016nesterovs}. The implemented Scaled YOLOv4 model is trained over a total of 47,224 batches of 8 images each.
        
\section{Image Analysis and Clustering} \label{sec:img-analysis-cls}
    This section details how an image is analyzed to extract detected humans coordinates, then to proceed with coordinates mapping and clustering.
    
    \subsection{Pinhole camera model}
        Due to the low-altitude wide-angle captured images in the VisDrone2019 dataset, close objects to the camera would appear larger/longer than further ones. To correctly represent this optical phenomenon, we leverage the pinhole camera model in the following analysis \cite{Sturm2014}. The latter is commonly used in computer vision to mimic the geometrical projection of light on the image plane of the camera.
        
    \subsection{Relation between the coordinates systems}
        In order to efficiently design a mapping function, two {orthogonal} coordinates systems are established, the image {plane} coordinates system {$(\bar{x},\bar{y})$} that localizes individuals' projection within the image plane of the camera and a real-world coordinates system $(\hat{x}, \hat{y}, {\hat{z}})$ that localizes individuals at a distance from the camera lens, illustrated in Fig. \ref{fig:pinhole-coords}. The ground is assumed perfectly flat, thus, any individual Z-coordinate is considered null in the real-world coordinates system.
        
        \Figure[t](topskip=0pt, botskip=0pt, midskip=0pt)[width=0.99\linewidth]{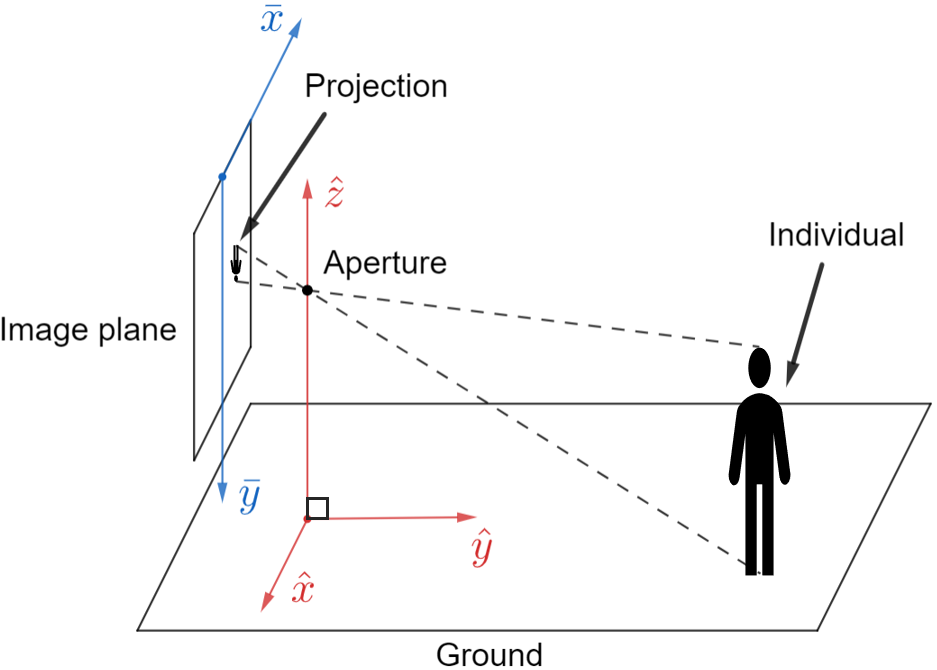}{Pinhole camera model illustration and relation between the image plane and real-world coordinates systems.\label{fig:pinhole-coords}}
        
        Any detected individual is delimited by a bounding box in the image plane. Bounding boxes are described using specific annotations. Indeed, the detection model outputs a vector for each bounding box presenting multiple features, including the bounding box location $(x_\text{left},\ y_\text{top})$, width $w$, and height $h$, in pixels, as shown in Fig. \ref{fig:image-coords}. For more accurate localization, we define new coordinates as follows:
        \begin{subequations}
            \begin{align}
                x_\text{center} &= x_\text{left} + \frac{w}{2} \\
                y_\text{bottom} &= y_\text{top} - h.
            \end{align}
        \end{subequations}
        
        \Figure[t](topskip=0pt, botskip=0pt, midskip=0pt)[width=0.99\linewidth]{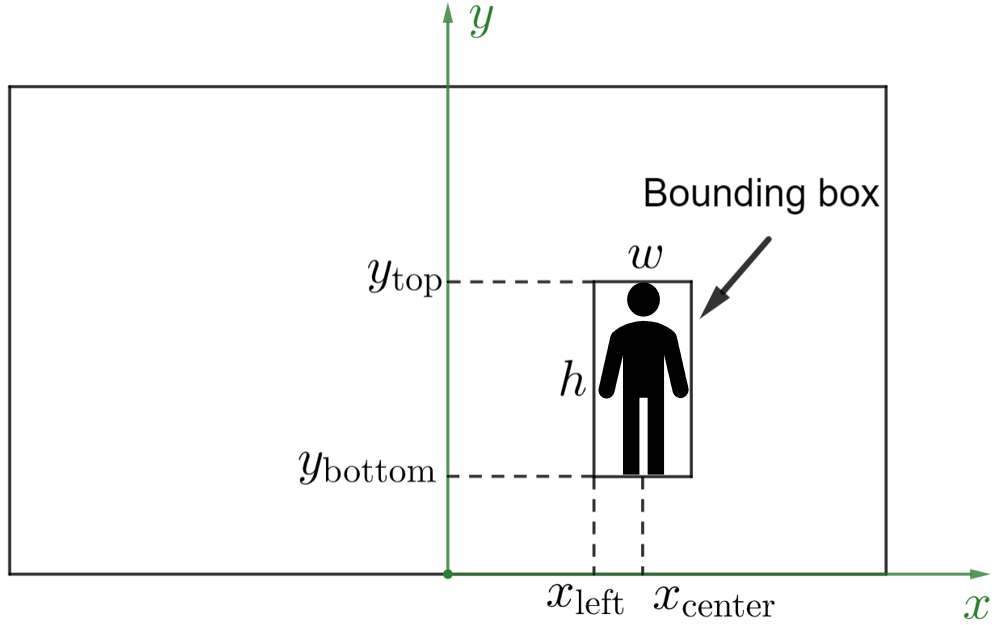}{Bounding box definition in the image coordinates system.\label{fig:image-coords}}
        
    \subsection{Bounding box correction}
        As mentioned formerly, our intention is to leverage the object as well as its projection's dimensions in order to estimate its location in the real-world coordinates system. Nonetheless, the objects in our setting are humans, where knowledge about the height of each individual cannot be determined, given the variable heights between adults, children, men, and women.
        Furthermore, bounding boxes fit only around pixels deemed to be part of a human's features and do not take into consideration posture and stance, i.e., standing, crouching, or sitting. As such, two major assumptions are made:
        
        \begin{itemize}
            \item People captured in an image are assumed to be perfectly standing adults.
            \item The height of a perfectly standing adult, either a man or a woman, is approximated to $1.75$ m.
        \end{itemize}
        
        These assumptions may seem naive since stumbling upon a group of $1.75$ m tall standing adults is very unlikely. Accordingly, children and seated/crouching adults will be perceived further than their actual real-world location as their bounding boxes are significantly smaller than that of a 1.75 m tall standing adult.
        In order to attenuate this effect, we introduce a bounding box correction mechanism. The idea is to bring all bounding boxes of detected individuals into an average state. Assuming that the majority of detected individuals follow the previous assumptions, the outliers, i.e., not standing adults or children, will see their bounding boxes adjusted to their average, with consideration to the pinhole camera model.
        
        For a set of $L$ original bounding boxes $(y^{(i)}_{\text{bottom}}, h^{(i)})$, $\forall i=1,\ldots,L$, drawn inside an image, a second degree polynomial function can be used to model the average height of each bounding box as a function of its Y-coordinate in the image plane, as follows:
            \begin{equation}
            \label{bbc}
                h^{(i)} = \alpha_0 + \alpha_1y^{(i)}_\text{bottom} + \alpha_2(y^{(i)}_\text{bottom})^2 + \epsilon^{(i)} = g(y^{(i)}_\text{bottom}) + \epsilon^{(i)},
            \end{equation}
        where $\epsilon^{(i)}$ is the deviation of the ${i}^{th}$ bounding box size from the standard size and $g(\cdot)$ is a second degree polynomial function with parameters $\alpha_0$, $\alpha_1$, and $\alpha_2$.
        
        In order to determine these parameters, we formulate the associated regression problem with the following least-square objective function:
        \begin{equation}
            \min_{\alpha_0, \alpha_1, \alpha_2} \sum_{i=1}^{L} (h^{(i)} - g(y^{(i)}_\text{bottom}))^2.
        \end{equation}
        Subsequently, $\alpha_0$, $\alpha_1$, and $\alpha_2$ are obtained by solving the equality of corresponding partial derivatives to 0. Eventually, the heights of bounding boxes are corrected by subtracting the deviation from their original heights as follows:
        \begin{equation}
            h_c^{(i)} = h^{(i)} - \epsilon^{(i)} = g(y^{(i)}_\text{bottom}),
        \end{equation}
        where $h_c^{(i)}$ is the corrected height of the $i^{th}$ bounding box.

    \subsection{Coordinates mapping}
        In order to obtain the metric dimensions of the objects projected on the image plane of the camera, we multiply coordinates and dimensions by the pixel size of the camera sensor, denoted $p$. Hence, we obtain
        \begin{subequations}
            \begin{align}
                \bar{x}_{\rm center} &= p \times x_\text{center} \\
                \bar{y}_{\rm bottom}&=p \times y_{\rm bottom}\\
                \bar{h} &= p \times h_c.
            \end{align}
        \end{subequations}
        
        % \begin{figure*}[t]
        %     \centering
        %     \subfloat[Projection on plane $(\hat{y}, \hat{z})$.]{\includegraphics[width=0.4\linewidth]{Figs/Pinhole-proj-yz}%
        %     }
        %     \hfill
        %     \subfloat[Projection on plane $(\hat{x}, \hat{y})$.]{\includegraphics[width=0.5\linewidth]{Figs/Pinhole-proj-xy}%
        %     }
        %     \caption{Projection of the pinhole camera model illustration on real-world system planes.}
        %     \label{fig:pinhole-proj}
        % \end{figure*}
        
        \begin{figure}[t]
            \centering
            \subfloat[Projection on plane $(\hat{y}, \hat{z})$.]{\includegraphics[width=0.88\linewidth]{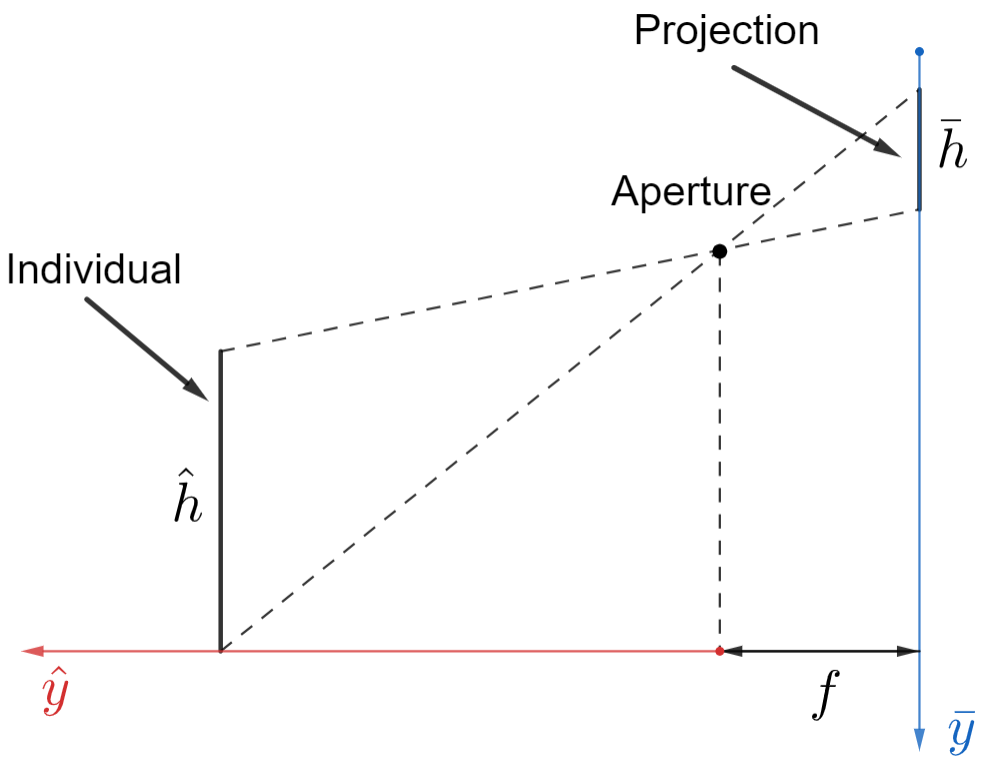}%
            }
            \vfill
            \subfloat[Projection on plane $(\hat{x}, \hat{y})$.]{\includegraphics[width=0.99\linewidth]{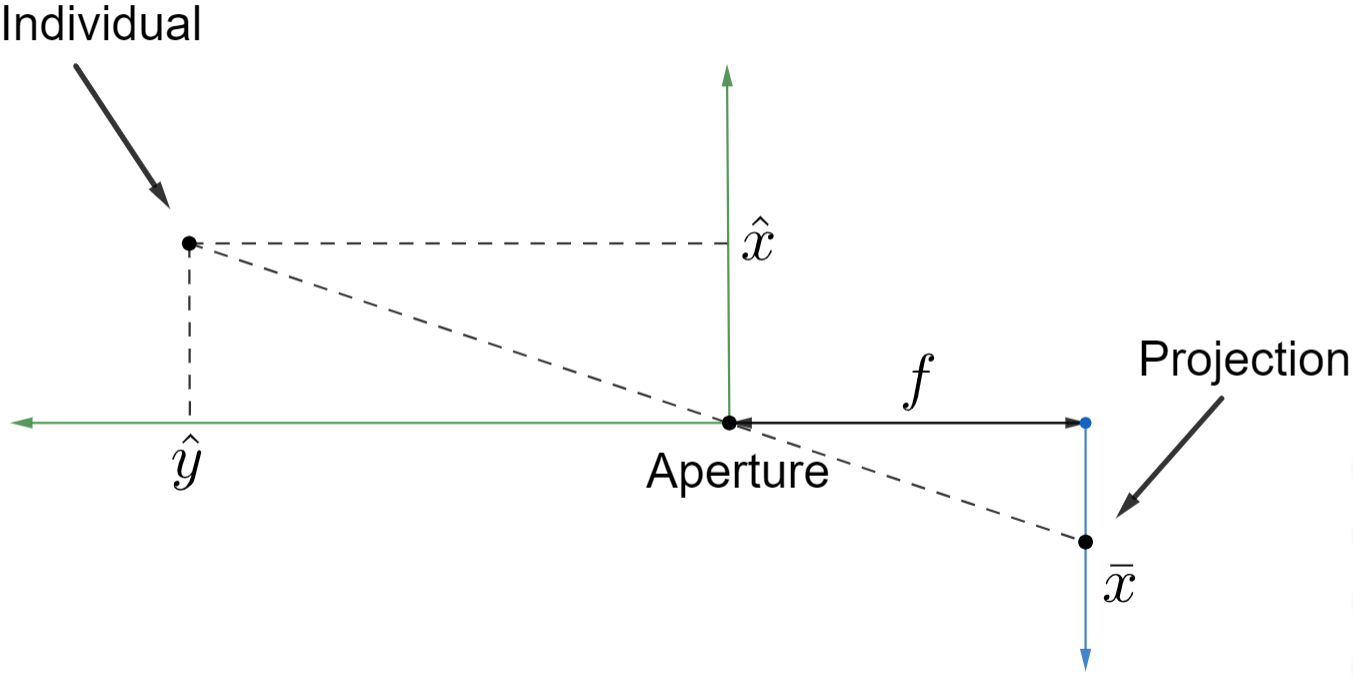}%
            }
            \caption{Projection of the pinhole camera model illustration on real-world system planes.}
            \label{fig:pinhole-proj}
        \end{figure}

        For coordinates mapping purposes, we assume that the camera axis is perfectly parallel to the ground plane. Also, let $f$ be the focal length of the camera, i.e., distance between aperture and image plane, and $(\hat{x}_{\rm center},\hat{y}_{\rm bottom})$ the estimated real-world ground coordinates of an individual in meters. Using different projection perspectives of the pinhole camera model, as illustrated in Fig. \ref{fig:pinhole-proj}, the mapping function is designed using the homothety rules, as follows:
        \begin{subequations}
            \begin{align}
            \hat{y} &= \frac{\hat{h} f}{\bar{h}} \label{eq:y-hat} \\
            \hat{x} &= \frac{\bar{x}_{\rm center}\ \hat{y}}{f} = \bar{x}_{\rm center}\frac{\hat{h}}{\bar{h}},
            \end{align}
        \end{subequations}
        where $\hat{h}$ is the height of the real-world object, assumed $\hat{h}=1.75$ m.
        
    \subsection{Density-based clustering}
        Given the randomness of the distribution of people captured by the UAV camera, partition-based and hierarchical clustering techniques are not adequate to identify groups with dense numbers of individuals. In contrast, density-based clustering is known to be efficient in recognizing randomly shaped clusters and in dealing with outliers. Hence, we opt here for density-based clustering to identify situations where social distancing is potentially not respected. Specifically, we use the popular Density-based Spatial Clustering of Applications with Noise (DBSCAN) algorithm for its intuitive and simple approach \cite{10.5555/3001460.3001507}.
        
\section{UAV trajectory optimization} \label{sec:uav-traj-opt}
    Within our crowd surveillance framework, we assume that the UAV acts in two phases. In the first, it flies at a relatively high altitude to detect people, determine their locations, and cluster them with respect to social distancing restrictions (Sections III and IV). In the second phase, it flies closer to the clusters with high risk of people closeness in order to verify, for instance, that mask wearing is in effect (Section V). The UAV trajectory for the second phase should be carefully designed in order to tradeoff between UAV energy consumption, i.e., shorter flying distance, and cluster visiting prioritization, i.e., clusters with higher risks should be visited first.
    Moreover, how to inspect each cluster (through hovering or slow flying) is a challenging task as it also needs to tradeoff between energy consumption and efficacy of inspection.
    Given the complexity of the aforementioned problem, the entire process is divided to two separate sub-problems. In the first, we optimize the main UAV trajectory, i.e., to fly from a cluster to another, while in the second, we design an efficient cluster inspection path.
    
    \subsection{Main trajectory optimization}
        The goal of the main trajectory design problem is to follow an energy-efficient path to visit all clusters and return to its initial location, while prioritizing high-risk clusters over low-risk ones. 
        For the sake of simplicity, we assume that the aerial environment is obstacle-free, and that the UAV flies sufficiently high to avoid collisions with people or objects. 
        This problem can be modeled using a complete directed graph $\mathcal{G} = (\mathcal{C}, \mathcal{E}, \mathbf{F})$ where $\mathcal{C} = \{0, \ldots, |\mathcal{C}|\}$ is the set of clusters, with $|\mathcal{C}|$ is the cardinality of $\mathcal{C}$ and cluster $0$ identifies the UAV's initial location. $\mathcal{E}$ is the set of edges connecting the clusters, and $\mathbf{F}= [f_{ij}]_{|\mathcal{C}| \times |\mathcal{C}|}$ is the cost matrix of the edges, where $f_{ij}$ corresponds to the cost of directed edge $(i,j)$. 
        Inspired by the Dantzig–Fulkerson–Johnson formulation \cite{tspform}, our problem can be written as
        
        \begin{subequations}\label{eq:opt-main}
            \begin{align}
                \min_{\mathbf{B}} \quad & \sum_{i = 0}^{|\mathcal{C}|-1}\sum_{j \neq i,j = 0}^{|\mathcal{C}|-1} b_{ij} f_{ij} \tag{\ref{eq:opt-main}}\\
                \text{s.t.} \quad & \sum_{\substack{i = 0\\ i \neq j}}^{|\mathcal{C}|-1} b_{ij} = 1 \label{eq:opt-const-a}\\
                \quad & \sum_{\substack{j = 0\\ j \neq i}}^{|\mathcal{C}|-1} b_{ij} = 1 \label{eq:opt-const-b}\\
                \quad & \sum_{i = 0}^{|\mathcal{C}|-1}\sum_{\substack{j = 0\\ j \neq i}}^{|\mathcal{C}|-1} b_{ij} = |\mathcal{C}|, \label{eq:opt-const-c}
            \end{align}
        \end{subequations}
        
        where $\mathbf{B} = [b_{ij}]_{|\mathcal{C}| \times |\mathcal{C}|}$ is a binary decision matrix that maps edges into the UAV trajectory, defined with
        
        \begin{equation}
            b_{ij} =
            \begin{cases}
                1, & \text{if path goes from cluster } i \text{ to cluster } j\\
                0, & \text{otherwise}.
            \end{cases}
        \end{equation}
        
        Constraints \eqref{eq:opt-const-a} and \eqref{eq:opt-const-b} guarantee that all clusters are visited only once, while \eqref{eq:opt-const-c} ensures that a single tour covers all clusters. Clearly, problem (\ref{eq:opt-main}) can be assimilated to a traveling salesman problem (TSP), which is known to be NP-hard. Consequently, by reduction, this problem is also NP-hard.
        
        Since our goal trades off between energy consumption and cluster priority adherence, we define the cost function of an edge as the weighted sum of two functions as follows:
        
        \begin{equation}
            f_{ij} = \alpha f^{p}_{ij} + (1 - \alpha) f^{e}_{ij},
        \end{equation}
        
        where $f^{p}_{ij}$ and $f^{e}_{ij}$ are the priority cost and energy consumption cost of edge $(i, j)$, respectively, and $\alpha \in [0, 1]$ is the weight.
            
        Assuming the UAV's velocity is uniform from one node to another, its energy consumption is linearly related to the traveled distance. Thus, the energy cost of an edge $(i, j)$ is equivalent to the distance between vertices $i$ and $j$. Subsequently, the energy consumption cost of edge $(i, j)$ can be defined by
        
        \begin{equation}
            f^{e}_{ij} = ||\mathbf{c}_j - \mathbf{c}_i||_2
        \end{equation}
        
        where $\mathbf{c}_i$ and $\mathbf{c}_j$ are the coordinates of clusters $i$ and $j$, respectively, defined as the barycenters of their clusters, and $||\cdot ||$ is the Euclidean norm. Moreover, the priority cost of an edge $(i, j)$ is defined as
        
        \begin{equation} \label{eq:priority-cost}
            f^p_{ij} =
            \begin{cases}
                \max(0,\ \lambda_j - \lambda_i), & \text{if } i \neq 0\\
                    0, & \text{otherwise}
            \end{cases}
        \end{equation}
        
        where $\lambda_i$ and $\lambda_j$ are the infection risk scores of clusters $i$ and $j$, respectively, where the infection risk of a cluster $i$ is given by
        
        \begin{equation}
            \lambda_i = \frac{2}{N_i (N_i-1)}  { \sum \limits_{l=1}^{N_i} \sum \limits_{\substack{k=l+1 \\ k > l}}^{N_i} \mathbbm{1}_{\{d_{kl} < 2\}}}+ N_i,
        \end{equation}
        
        where $k$ and $l$ are the indexes of detected individuals in cluster $i$, $d_{kl}$ the distance in meters separating them, and $N_i$ is the number of individuals in cluster $i$. Finally, $\mathbbm{1}_{x}$ assigns value 1 when condition $x$ is satisfied. To be noted that $\lambda_i$ is the sum of two components, the first is the social distancing disrespect ratio, while the second is the size of the cluster.
        
        Several approaches can be used to solve problem (\ref{eq:opt-main}). Intuitively, one would explore all possible combination in the graph $\mathcal{G}$ to obtain the optimal solution. This exhaustive search may work optimally for small-sized systems, however, it scales poorly with the number of clusters as $\mathcal{O}(|\mathcal{C}|!)$. For instance, with only 10 clusters more than 3 millions trajectory combinations need to be evaluated, which is very resource and time consuming, two factors that are scarce in crowd surveillance operations. Nevertheless, it is considered as a benchmark (for small-sized systems) in the next Section. Alternatively, we opt for heuristic and metaheursitic approaches as follows:
        
        \begin{itemize}
            \item \textbf{2-Opt \cite{10.2307/167074}} is a lightweight iterative parameter-free TSP algorithm that consists of systematically swapping paths between vertices until a certain optimum is reached. We summarize our implementation of 2-Opt to solve problem (\ref{eq:opt-main}) in Algorithm \ref{alg:2-opt}.
            
            \begin{algorithm}[t]
                \caption{2-Opt Algorithm}
                \label{alg:2-opt}
                \SetKwInOut{Input}{Input}
                \SetKwInOut{Output}{Output}
                \SetKwRepeat{Do}{do}{while}
                \SetKw{KwBy}{by}
                \Input{Set of clusters $\mathcal{C}$ and corresponding cost matrix $\mathbf{F}$.}
                \Output{Optimized UAV trajectory $t^{*}$.}
                \BlankLine
                $t \gets$ random trajectory \;
                $improved \gets True$\;
                \While{$improved$}{
                    $improved \gets False$\;
                    \For{$i\gets 0$ \KwTo $|\mathcal{C}| - 2$ \KwBy $1$}{
                        \For{$j\gets i+2$ \KwTo $|\mathcal{C}|$ \KwBy $1$}{
                            \If{$f_{i(i+1)}+f_{j(j+1)} > f_{ij}+f_{(i+1)(j+1)}$}{
                                Swap edges $i$ and $j$ of trajectory $t$\;
                                $improved \gets True$\;
                            }
                        }
                    }
                }
                $t^* \gets t$\;
            \end{algorithm}
        \end{itemize}
        
        \begin{itemize}
            \item \textbf{Genetic Algorithm (GA) \cite{10.5555/534133}} reflects the process of natural selection where the fittest solutions are selected for breeding in order to produce better offspring for the upcoming generation. It evaluates full trajectories using the fitness function. Given a trajectory or route $r$, the fitness value $\phi$ is given by
            
            \begin{equation} \label{eq:ga-fitness}
                \phi(r) = \frac{1}{\sum_{(i, j) \in \mathcal{E}} b_{ij} f_{ij}}.
            \end{equation}
            
            For problem (\ref{eq:opt-main}), the GA algorithm is presented in Algorithm \ref{alg:genetic-alg}. GA population $\mathcal{P}$ is a set of UAV trajectories. 
            A gene is a cluster to visit. An individual is a UAV trajectory satisfying constraints (\ref{eq:opt-const-a})-(\ref{eq:opt-const-c}). 
            The parents are combined solutions. The mating pool is a collection of parents that creates the next generation.
            Mutations introduce variations in the population by randomly swapping clusters for a trajectory. Finally, elitism carries the best individuals to the next generation. The rank selection method is preferred over the roulette wheel approach to minimize the election of defective solutions for the mating pool. 
            
            Also, given the order constraint of the TSP, the ordered-crossover operator is used to generate offspring.
            
            \begin{algorithm}[t]
                \caption{Genetic Algorithm}
                \label{alg:genetic-alg}
                \SetKwInOut{Input}{Input}
                \SetKwInOut{Output}{Output}
                \SetKw{KwBy}{by}
                \Input{Set of clusters $\mathcal{C}$ and corresponding cost matrix $\mathbf{F}$.}
                \Output{Optimized UAV trajectory $t^{*}$.}
                \BlankLine
                $\mathcal{P} \gets$ Set of random trajectories (population)\;
                \tcp{$N_g$: number of generations}
                \For{$i\gets1$ \KwTo $N_g$ \KwBy $1$}{
                    Initialize an empty set $\widetilde{\mathcal{P}}$\;
                    Calculate fitness of individuals in $\mathcal{P}$ using \eqref{eq:ga-fitness}\;
                    Add the two fittest solutions of  $\mathcal{P}$ to the new set $\widetilde{\mathcal{P}}$\;
                    Select mating pool using rank selection ($\mathcal{N} \subset \mathcal{P}$)\;
                    Create offspring from mating pool $\mathcal{N}$ using ordered crossover\;
                    Use swap mutation on offspring\;
                    Add mutated offspring to $\widetilde{\mathcal{P}}$\;
                    $\mathcal{P} \gets \widetilde{\mathcal{P}}$\;
                }
                $t^* \gets$ solution with best fitness from $\mathcal{P}$\;
            \end{algorithm}
            
            \item \textbf{Ant Colony Optimization (ACO) \cite{Colorni1992DistributedOB}} emulates the swarm behavior of ants when looking for the shortest route from their shelter to a food source. In fact, randomly placed artificial ants inside a graph gradually advance through a Hamiltonian cycle using a probabilistic approach based on the pheromone intensity of the edge as well as the visibility of the target node. Such process is simulated through multiple ant colonies until a convergence criteria is met.
            ACO is similar to GA in the way it considers weights of edges as rewards rather than penalties. Thus, we define its visibility matrix as $\mathbf{H} = [\eta_{ij}] = [1/f_{ij}]_{|\mathcal{C}| \times |\mathcal{C}|}$.
            We implement ACO to solve our problem as described in Algorithm \ref{alg:aco}.
            
            \begin{algorithm}[t]
                \caption{ACO Algorithm}
                \label{alg:aco}
                \SetKwInOut{Input}{Input}
                \SetKwInOut{Output}{Output}
                \SetKw{KwBy}{by}
                \Input{
                    Set of clusters $\mathcal{C}$, corresponding cost matrix $\mathbf{F}$, colony size $s$, information elicitation factor $\delta$, $\beta$, pheromone intensity $Q$, and pheromone evaporation coefficient $\rho$.
                }
                \Output{Optimized UAV trajectory $t^{*}$.}
                \BlankLine
                $\mathbf{H} \gets [1/f_{ij}]_{|\mathcal{C}| \times |\mathcal{C}|}$\;
                Initialize pheromone matrix $\mathbf{T}$\;
                \For{$i\gets1$ \KwTo $N$ \KwBy $1$}{
                    Place a set $\mathcal{K}$ of $s$ ants in cluster 0\;
                    Calculate node selection probability matrix\;
                    Advance ants through a Hamiltonian cycle using predetermined probabilities\;
                    Update pheromone matrix $\mathbf{T}$\;
                }
                $t^* \gets$ best route found by the last colony $\mathcal{K}$\;
            \end{algorithm}
        \end{itemize}
        
    \subsection{Cluster inspection path design}
        Once clusters and formed and main UAV trajectory is designed, the UAV has to adopt an inspection strategy when getting closer to clusters. Since individuals may be standing in different angles with respect to the UAV's perspective, we opt for a circular inspection path that would capture individuals from the front. Nevertheless, a safety distance must be kept to avoid any harm to individuals or to the UAV. Hence, we design the inspection path in two steps, namely convex hull design and safety distance design.
        
        Since the size of a cluster is small, we propose to use the Jarvis March algorithm, which determines the convex hull surrounding a cluster's individuals \cite{JARVIS197318}. The latter is time-efficient, achieving a complexity of $\mathcal{O}(N_i \eta_i)$, where $\eta_i$ is the number of points of the convex hull, linked to cluster $i$. The convex hull approach implies that the UAV has to fly dangerously and directly over some of the individuals in a cluster (the ones being included in the convex hull). This might compromise both safety and the quality of inspection for those individuals. Consequently, we expend the convex hull method with a safety constraint as follows.
        
        First, let $\mathcal{W} = \{\omega_1, \omega_2, \dots, \omega_{|\mathcal{W}|}\}$ be the set of convex hull points related to a cluster, and the closed ordered chain of points $\mathcal{W'}=\{\omega_1, \omega_2, \dots, \omega_{|\mathcal{W}|}, \omega_1\}$ is oriented clockwise. Consequently, the following convex hull property holds:
        
        \begin{equation} \label{eq:safe-dist-barycenter}
                (\overrightarrow{{\omega_i \omega_j}} \land \overrightarrow{{\omega_io}}) \cdot {{\hat{z}}} < 0, \; \forall \; 1 \leq i < j \leq |\mathcal{W}|,
        \end{equation}
        
        where $o$ is the barycenter of the shape constructed using $\mathcal{W'}$'s points, while $\land$ and $\cdot$ are the cross-product and dot-product of two vectors, respectively. Property (\ref{eq:safe-dist-barycenter}) allows to identify the outbound of a cluster, from one convex hull edge perspective.
        Subsequently, we propose to geometrically translate each convex hull edge away from the cluster by a safety distance $d_S$ along one of its normal vectors ${\overrightarrow{v}}$, i.e.,
        
        \begin{equation} \label{eq:safe-dist-normal}
        (\overrightarrow{{\omega_i\omega_j}} \land {d_S \overrightarrow{v}})\cdot {{\hat{z}}} > 0, \; \forall \; 1 \leq i < j \leq |\mathcal{W}|.
        \end{equation}
        
        Afterwards, to complete the novel path design, we fill the gaps between the translated convex hull edges using arcs with centers as the points in $\mathcal{W}$ and radius equal to $d_S$. The use of arcs ensures that safety distance is respected along the whole inspection path.

%\newpage
        
\section{Results and Discussion} \label{sec:res-dis}
     \subsection{Human detection}
    	The trained scaled YOLOv4 model is evaluated on 548 pre-processed images of the validation set of VisDrone2019 using the AP performance metric defined as follows:
    	
    	\begin{equation}
    		AP = \int_0^1 \nu(r)dr
    	\end{equation}
    	
    	where $\nu(r)$ is the precision-recall curve, obtained using the precision ($\gamma_p$) and recall ($\gamma_r$) metrics. The latter are defined as    	
    	
    	\begin{subequations}
        	\begin{align}
    		    \gamma_p &= \frac{T_p}{T_p + F_p}, \\
    			\gamma_r &= \frac{T_p}{T_p + F_n},
    		\end{align}
    	\end{subequations}
    	
    	where $T_p$ is the number of true positives, $F_p$ of false positives, and $F_n$ of false negatives. These parameters are calculated by comparing the Intersection over Union (IoU) measure of each predicted bounding box with a threshold set to 0.5.
    	The IoU of a predicted bounding box is given by
    	
    	\begin{equation}
    	    IoU = \frac{A_o}{A_u},
    	\end{equation}
    	
    	where $A_o$ is the area of overlap between the ground truth (in annotated images) and the prediction, and $A_u$ is the area of union of the ground truth and the prediction.
    	Moreover, the average IoU of predicted bounding boxes is calculated to assess how well the model fits boxes around individuals.
    	
    	\setcounter{table}{2}
        \begin{table}[t]
            \centering
        \caption{Performance of the proposed human detection model.}
            \begin{tabular}{lc}
                \hline
                {AP} & 65\% \\
                \hline
                {Precision} &  71\%\\
                \hline
                {Recall} & 61\% \\
                \hline
            {Average IoU} & 51.68\% \\
                \hline
            \end{tabular}
            \label{tab:det-eval}
        \end{table}
    	
    	In Table \ref{tab:det-eval}, we present the performances of our human detection model. First, it achieves AP of 65\%, which is caused by a degraded recall of 61\%. The latter means that only 61\% of individuals in the evaluation set were identified. This degraded performance is issued from the difficulty encountered by the model to detect partially or highly occluded individuals, as shown in Fig. \ref{fig:det-inf} below.  
    	Besides, the model fails to accurately replicate annotated bounding boxes as the average IoU is significantly low (around 52\%). Such a defect can be caused by the dataset's annotation errors, which may disturb both the model's training and evaluation processes. For instance, we notice in Fig. \ref{fig:det-inf} that our model correctly fitted some bounding boxes around individuals, as opposed to the erred ground truth, while in other occasions, it failed to do so. 
    	
    	Clearly, the dataset's heterogeneity, specifically the variation in scale and perspective of captured images, prevents the model from generalizing efficiently. Nevertheless, when compared to state-of-the-art methods in Table \ref{tab:det-comp}, we see that our model outperforms the approaches of \cite{tang2020penet,albaba2020synet,yu2020resolving}, while achieves the same AP as the CenterNet model \cite{pailla2019object}. Finally, our model is more interesting than CenterNet since it trains fast, while the CenterNet is known for its slow convergence and time-inefficiency as demonstrated in \cite[Table 9]{DBLP:journals/corr/abs-2004-10934}.

        % \begin{figure*}[t]
        %     \centering
        %     \subfloat[Image 0000022\_00500\_d\_0000005.]{\includegraphics[width=0.45\linewidth]{Figs/Inference1.pdf}%
        %     }
        %     \hfill
        %     \subfloat[Image 0000086\_00000\_d\_0000001.]{\includegraphics[width=0.45\linewidth]{Figs/Inference2.pdf}%
        %     }
        %     \caption{Inference results on a sample from VisDrone2019 images.}
        %     \label{fig:det-inf}
        % \end{figure*}
        
        \begin{figure}[t]
            \centering
            \subfloat[Image 0000022\_00500\_d\_0000005.]{\includegraphics[width=0.99\linewidth]{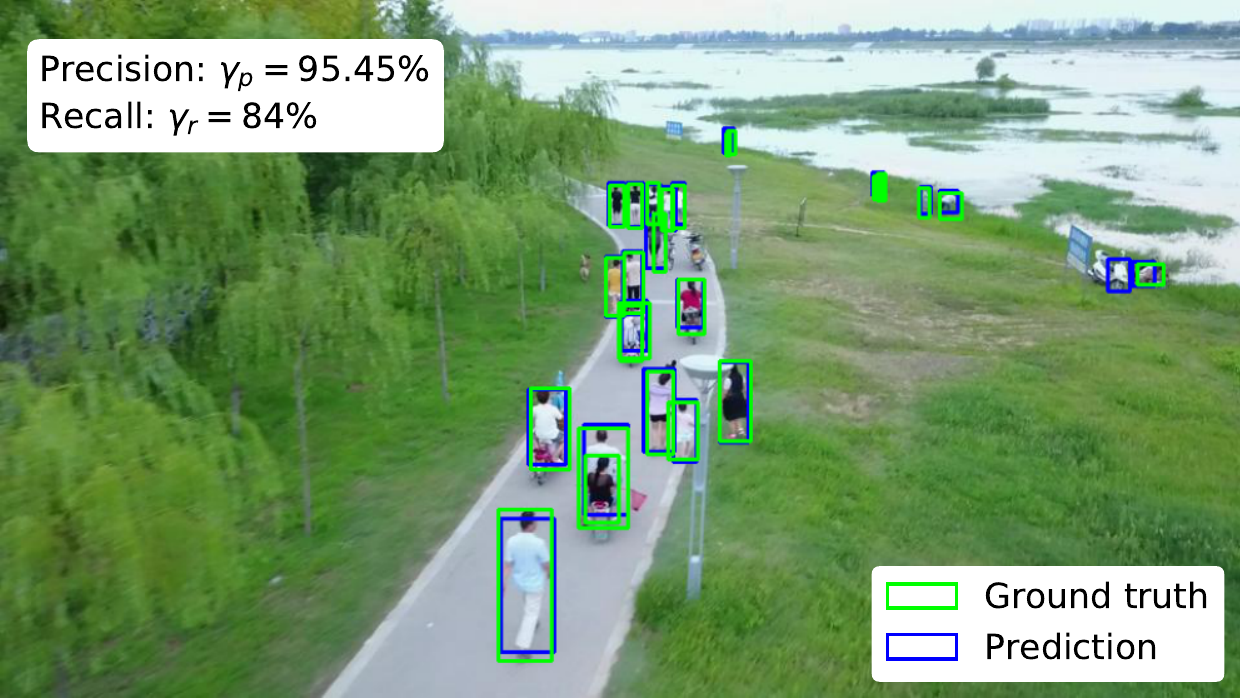}%
            }
            \vfill
            \subfloat[Image 0000086\_00000\_d\_0000001.]{\includegraphics[width=0.99\linewidth]{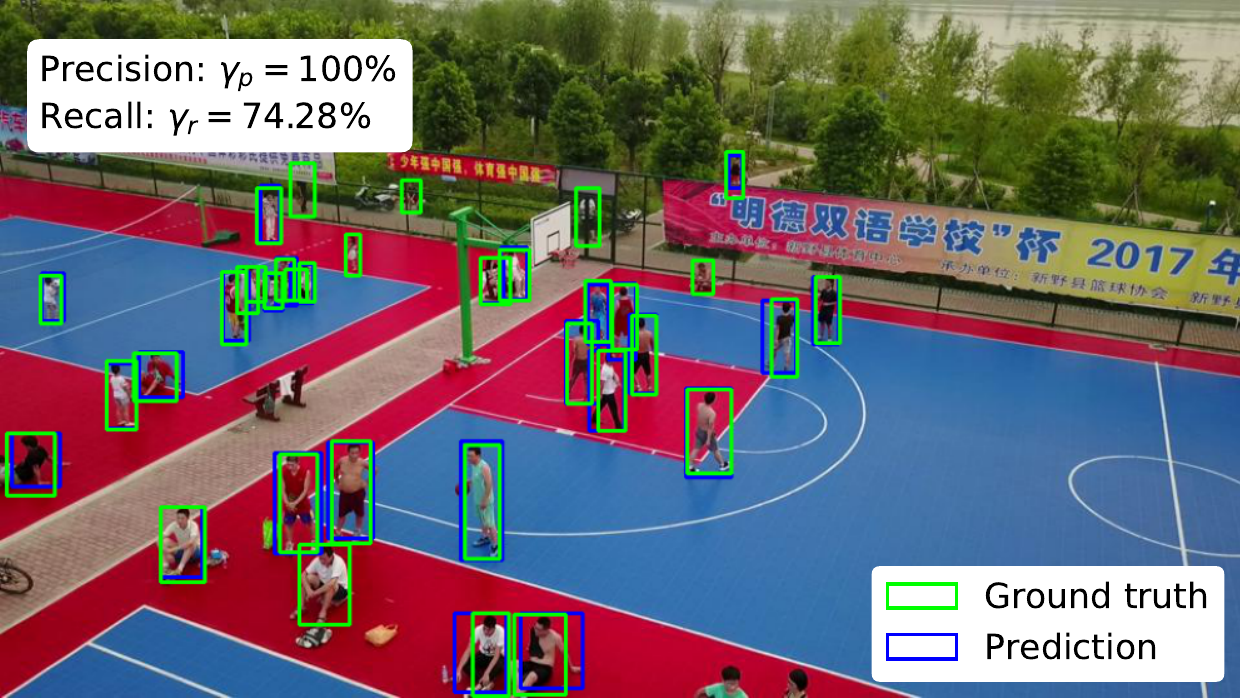}%
            }
            \caption{Inference results on a sample from VisDrone2019 images.}
            \label{fig:det-inf}
        \end{figure}
        
        \begin{table}[t]
            \centering
            \caption{Comparison of different human detection models.}
            \begin{tabular}{lc}
                \hline
                \textbf{Method} & \textbf{AP} \\
                \hline
                CenterNet \cite{pailla2019object} &  65\%\\
                % \rowcolor{yellow!50} CenterNet \cite{pailla2019object} &  65\%\\
                \hline
                PENet \cite{tang2020penet} & 41\% \\
                \hline
                SyNet \cite{albaba2020synet} & 43\%\\
                \hline
                DSHNet \cite{yu2020resolving} & 19.5\% \\
                \hline
                % \rowcolor{yellow!50} Our model & 65\% \\
                Our model & 65\% \\
                \hline
            \end{tabular}
            \label{tab:det-comp}
        \end{table}
        
    \subsection{Image analysis and clustering}
        {The evaluation of image analysis and clustering is split into three phases, namely bounding box correction, coordinates mapping, and individuals clustering.
        Due to the absence of a ground truth for assessment, these phases are evaluated based on the plain human observation.
        Moreover, in order to prevent error propagation from the human detection model, original annotations of bounding boxes are used for coordinates mapping and clustering. For the sake of clarity, the results are presented for the same VisDrone2019 sample image, which is shown in Fig. \ref{fig:det-inf}a.}
        
        Bounding box correction aims to bring original bounding boxes of detected individuals into an average state, under the assumption that the height of all individuals is around $1.75$ m. 
        Consequently, the bounding boxes of outliers, i.e., not  standing adults and  children, are averaged to their average value, with respect to the pinhole camera model.
        The results of this task are illustrated in Fig. \ref{fig:bbox-corr}, where green rectangles are the original bounding boxes and red rectangles the corrected ones. As it can be seen, the correction may consist on either reducing the height of the bounding box, e.g., in the case of adults, or on increasing its height, e.g., for children.
        
        \Figure[t](topskip=0pt, botskip=0pt, midskip=0pt)[width=0.99\linewidth]{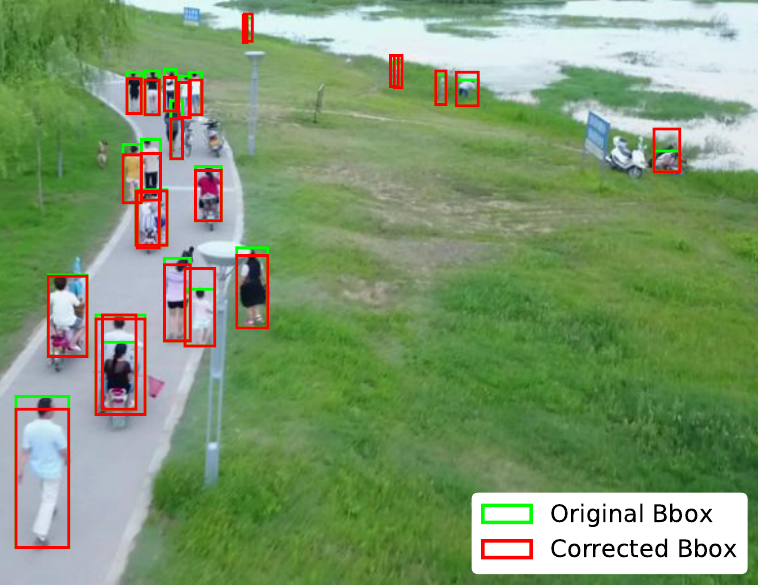}{Bounding box (Bbox) correction performed on VisDrone2019 image 0000022\_00500\_d\_0000005.\label{fig:bbox-corr}}
        
        Next, given the corrected bounding boxes, the coordinates mapping task is evaluated. The proposed mapping function relies on two camera parameters, namely the focal length of the lens $f$ and the pixel size of the sensor $p$. These parameters were not provided with the VisDrone2019 dataset and the AISKEYE team did not affirm that all images were captured using the same camera specifications. To accurately execute the coordinates mapping task, we focus at first on determining the best pair $(f,p)$ that delivers most accurate results. Specifically, we apply coordinates mapping for several combinations of $(f,p)$ values over a random set of images. Then, based on human observation, we decide on which combination that provides the most accurate coordinates mapping output with respect to the original image. We found that the most accurate parameters that give acceptable representations for the selected random set of images are $(f,p)=(10 \text{ mm}, 18 \; \mu \text{m} )$.
        Subsequently, coordinates mapping is executed with these parameters and the output is shown in Fig. \ref{fig:coords-map}. Indeed, the estimated locations of detected individuals are plotted in the real-world coordinates system as blue dots. 
        
        In Fig. \ref{fig:det-influence}, we investigate the impact of human detection errors (i.e., misplaced bounding boxes) on the coordinates mapping performance. To do so, we establish the relation between the height of the annotated bounding boxes $h$, height of estimated bounding boxes $\hat{h}$, and the Y-axis original and estimated coordinates $y$ and $\hat{y}$ (calculated using \eqref{eq:y-hat}).  
        As it can be seen, when $h$ is small, the gap in coordinates mapping $|y-\hat{y}|$ is very high for any height gap $|h-\hat{h}|$. For instance, if $h=40$ pixels (a typical value encountered with the VisDrone 2019 dataset) and the estimated bounding box is only 2 pixels away from the original one, the individual's location is shifted by 1 m, while a gap of 6 pixels in $|h-\hat{h}|$ causes a location estimation error of about 4 m. Although these values may seem small, they are significant when monitoring social distancing, which typically requires 2 m distance between individuals, in addition to leading to incorrect clustering.
        
        \Figure[t](topskip=0pt, botskip=0pt, midskip=0pt)[width=0.99\linewidth]{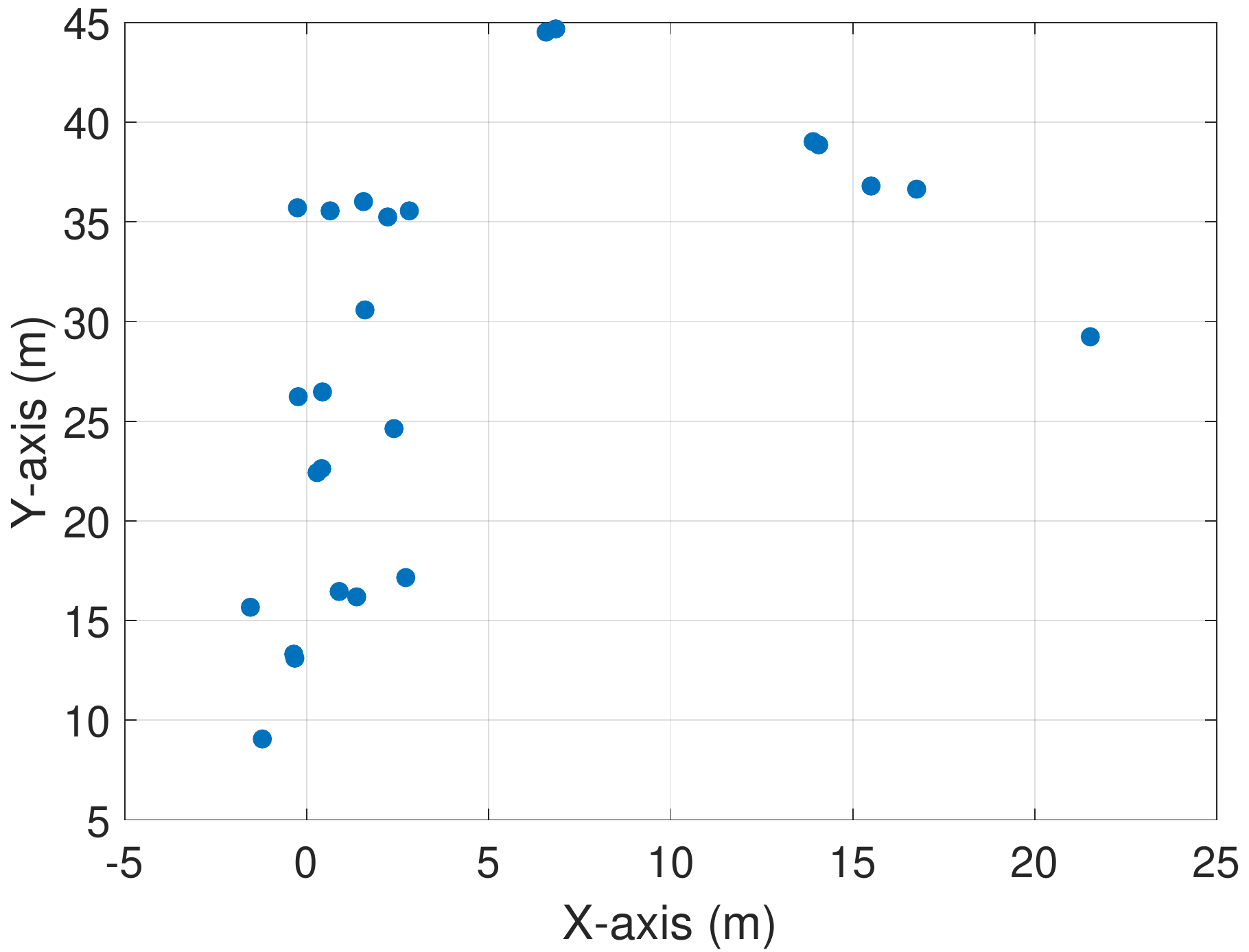}{Coordinates mapping applied on corrected bounding boxes of VisDrone2019 image 0000022\_00500\_d\_0000005.\label{fig:coords-map}}
        
        \Figure[t](topskip=0pt, botskip=0pt, midskip=0pt)[width=0.99\linewidth]{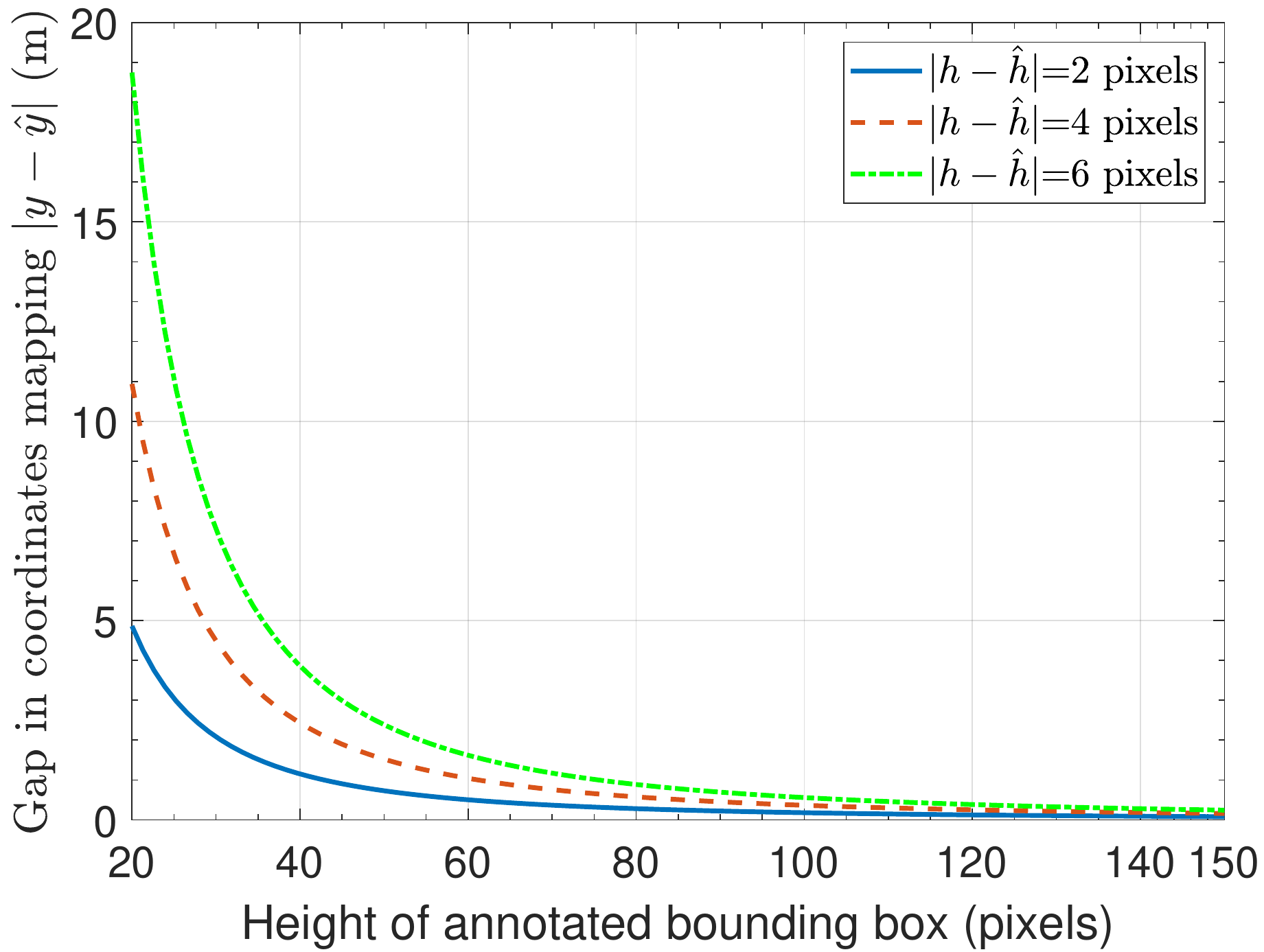}{Impact of bounding box estimation error on coordinates mapping.\label{fig:det-influence}}
        
        Following the coordinates mapping phase, DBSCAN algorithm is applied for clustering. We set the DBSCAN parameters as follows: The social distancing threshold $\epsilon = 2$ m, while the minimum number of individuals within a cluster is defined as $m = 3$. In other words, groups of a single or two people are considered as outliers. The reason behind it is that groups of two people are likely to be from the same household, and thus do not need to respect social distancing. As shown in Fig. \ref{fig:cls-test}, clusters were correctly identified, including children and seated individuals.
        
        \Figure[t](topskip=0pt, botskip=0pt, midskip=0pt)[width=0.99\linewidth]{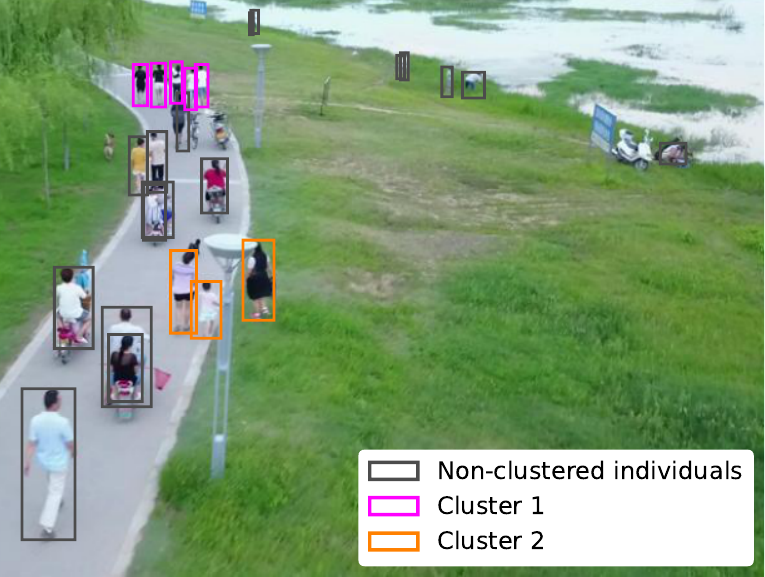}{Clustering of detected individuals on VisDrone2019 image 0000022\_00500\_d\_0000005.\label{fig:cls-test}}

        \begin{figure*}[t]
            \centering
            \subfloat[Total cost vs. number of clusters $(\alpha = 0)$.]{\includegraphics[width=0.45\linewidth]{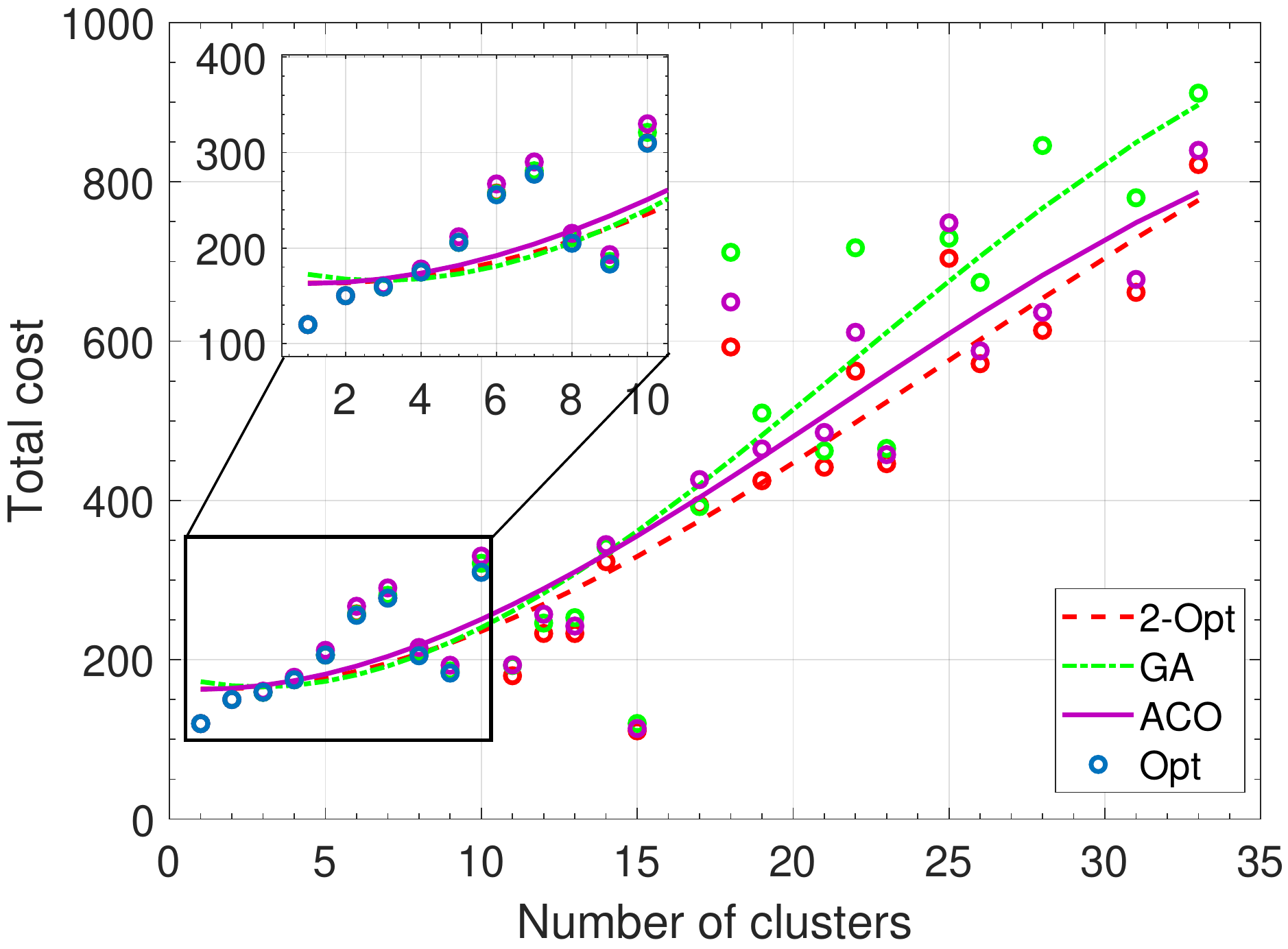}\label{fig:traj-opt-eval-alpha0}}%
            \hfil
            \subfloat[Total cost vs. number of clusters $(\alpha = 1)$.]{\includegraphics[width=0.45\linewidth]{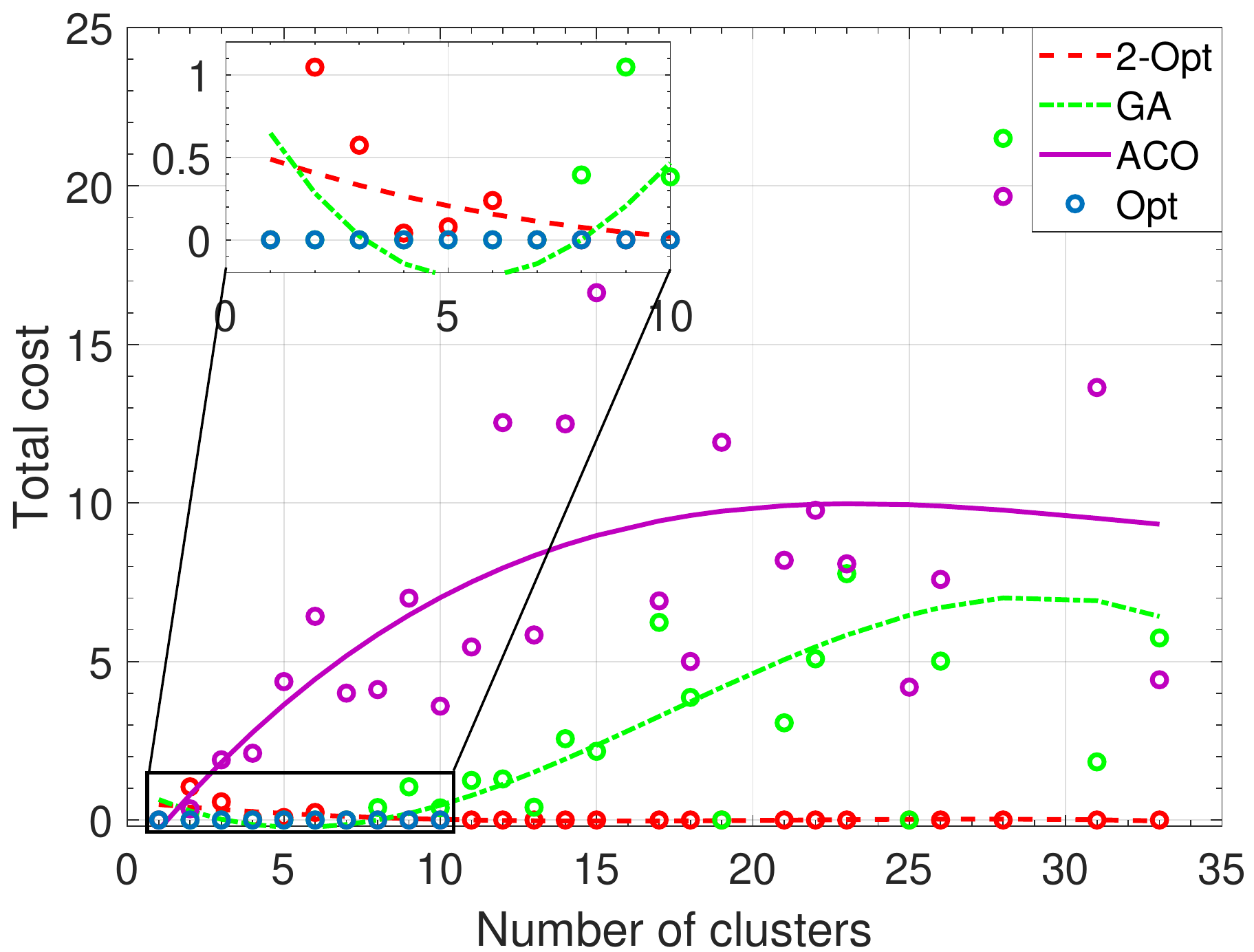}\label{fig:traj-opt-eval-alpha1}}%
            \vskip\baselineskip
            \subfloat[Total cost vs. number of clusters $(\alpha = 0.99)$.]{\includegraphics[width=0.45\linewidth]{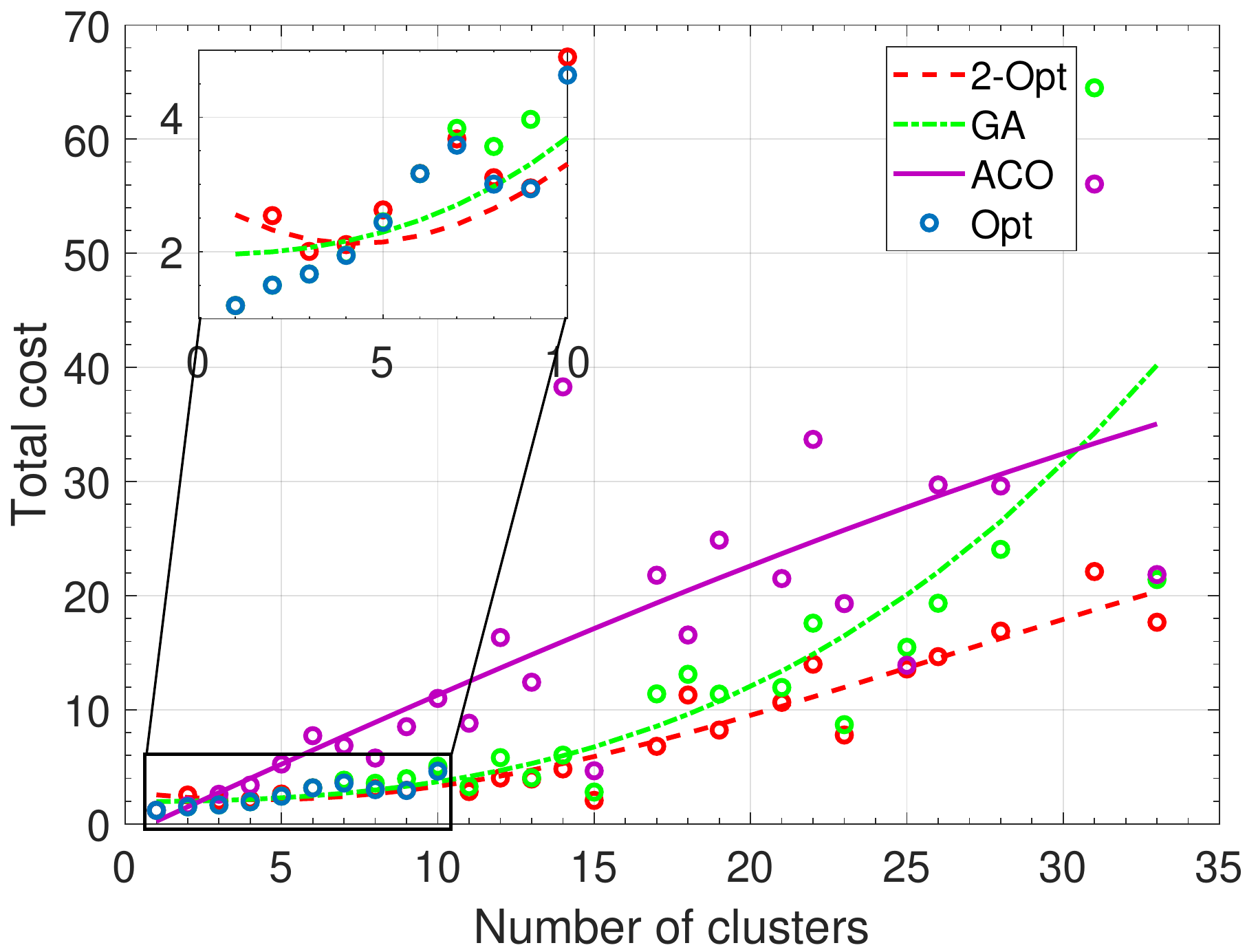}\label{fig:traj-opt-eval-alpha099}}%
            \hfil
            \subfloat[Execution time vs. number of clusters ($\alpha=0.99$).]{\includegraphics[width=0.45\linewidth]{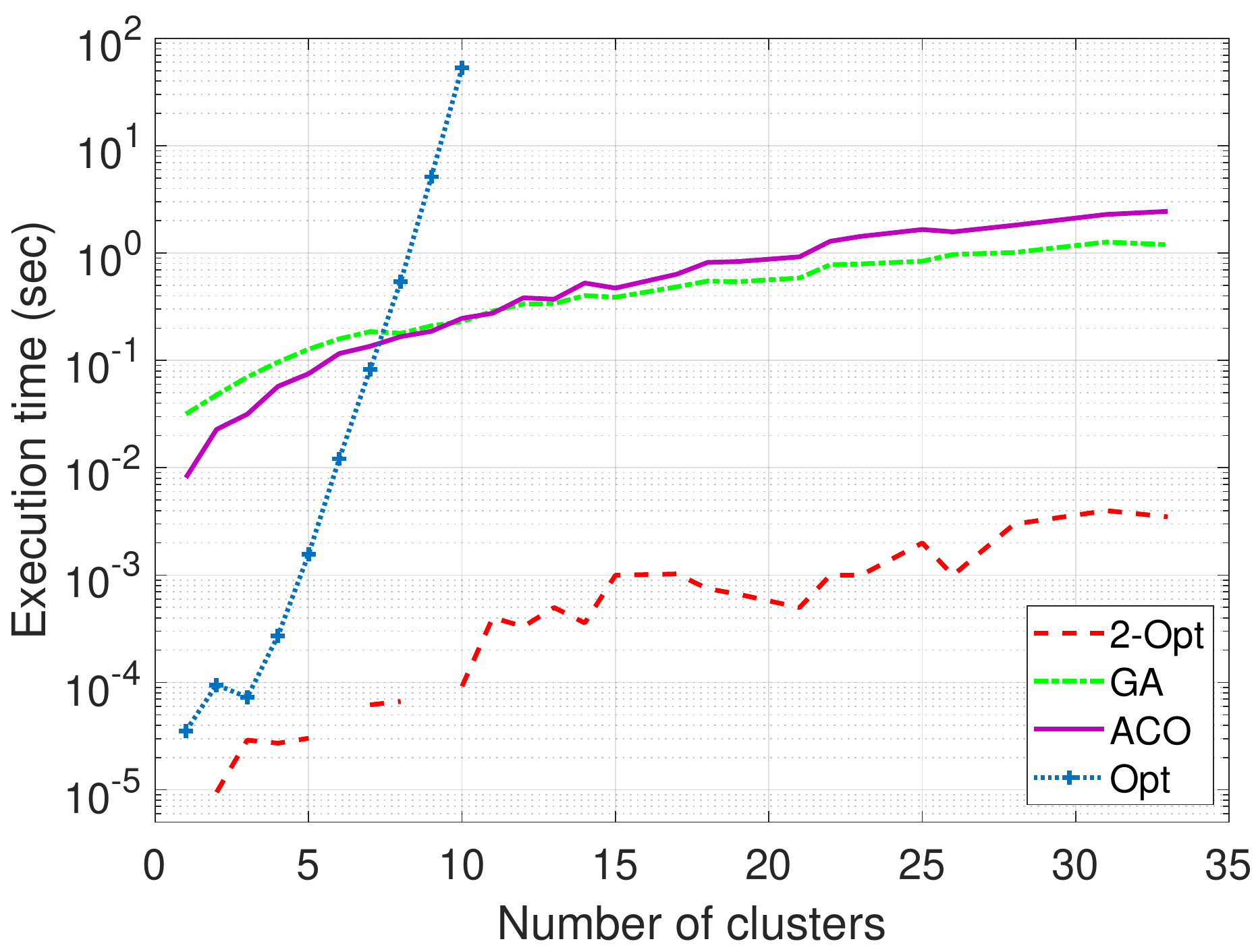}\label{fig:traj-opt-eval-exec-time}}%
            
            \caption{Performances of UAV trajectory planning algorithms (total cost and execution time) vs. number of clusters.}
            \label{fig:traj-opt-eval}
        \end{figure*}
        
    \subsection{UAV trajectory optimization}
        The UAV trajectory has two components, namely the main path and the cluster inspection path.
        
        For the main path optimization, we choose our baseline as the optimal solution, which is obtained through exhaustive search for small-scale scenarios (i.e., images with less than 10 clusters) due to its high time complexity. The latter is compared to the proposed heuristic and meta-heuristic based ones, i.e., based on Algorithms 1 (2-Opt), 2 (GA), and 3 (ACO). 
        We set the following parameters for GA and ACO:
        
        \begin{itemize}
            \item \textbf{GA:} For mutations, a random swap of nodes is implemented with rate 10\%. To maintain the evolutionary aspect of the algorithm, only top-two solutions are considered for elitism. Due to the variability in number of clusters identified in each image sample of the VisDrone2019 dataset, the population size is set to twice the number of clusters. This allows to avoid both over-population and under-population, improve convergence, and explore a larger search space.
        
            \item \textbf{ACO:} The colony size is set dynamically to cope with the heterogeneity of data. The remaining of its parameters are set as in Table \ref{tab:aco-params}.
        \end{itemize}
        
        \begin{table}[t]
            \centering
            \caption{ACO parameters.}
            \begin{tabular}{ll}
                \hline
                Colony size & $s = \lceil (|\mathcal{C}|-1)/2 \rceil$\\
                \hline
                Information elicitation factor & $\delta = 1$\\
                \hline
                Expected heuristic factor &  $\beta = 5$\\
                \hline
                Pheromone intensity & $Q = 10$\\      
                \hline
                Pheromone evaporation coefficient & $\rho = 0.5$\\
                \hline
            \end{tabular}
            \label{tab:aco-params}
        \end{table}
        
        In Fig. \ref{fig:traj-opt-eval}, we present the performances of the trajectory optimization algorithms in terms of average total cost \eqref{eq:opt-main} and execution time, as functions of the number of clusters $|\mathcal{C}|$, and for different cost weights $\alpha$. Due to the small number of images with high number of clusters, averaged results (circles) may not present a smooth trend. Hence, we opted to add the results of third order polynomial curve fitting (solid lines) to better show the trend with the increasing number of clusters.
        When $\alpha=0$ (Fig. \ref{fig:traj-opt-eval-alpha0}), the trajectory is optimized with regards to energy-efficiency only, reflected here through the traveled distance of the UAV. Obviously, the total cost increases rapidly with the number of clusters. This is expected since a higher $|\mathcal{C}|$ means that the UAV has to fly for longer distances to visit all clusters. Also, we notice that 2-Opt performs best compared to GA and ACO for any $|\mathcal{C}|$, and it achieves optimality for $|\mathcal{C}|<10$.
        When $\alpha=1$ (Fig. \ref{fig:traj-opt-eval-alpha1}), the UAV path is determined to adhere to the cluster priority order, i.e., prioritizing visiting clusters with high risk of social distancing violation.
        We notice here that the total cost increases slowly with $|\mathcal{C}|$, with preference for the 2-Opt based approach. However, 2-Opt is now sub-optimal for $|\mathcal{C}|<10$. This is predictable since 2-Opt is a heuristic approach that typically finds near-optimal solutions. In addition, we remark that the priority adherence cost is approximately two orders of magnitude smaller than that of the traveled distance (compared to Fig. \ref{fig:traj-opt-eval-alpha0}). Thus, to accurately balance between them in the cost function, we set $\alpha=0.99$ for the simulations of Fig. \ref{fig:traj-opt-eval-alpha099}. In this scenario, 2-Opt is near-optimal for $|\mathcal{C}|<10$ while it outperforms GA and ACO solutions for any $|\mathcal{C}|$ value. 
        Finally, the related execution times are evaluated in Fig \ref{fig:traj-opt-eval-exec-time}. In addition to its near-optimal performances, 2-Opt demonstrates the fastest execution time (in the range of micro and milliseconds), compared to exhaustive search, GA, and ACO. Hence, 2-Opt is seen as a practical solution for deployment in real-time environments.
        
        For cluster inspection, we illustrate in Fig. \ref{fig:cvx-hull-sim} the construct of the inspection trajectory around one cluster. As it can be seen, the Jarvis March algorithm designs a non-smooth curve (dashed red line) around the cluster. Due to security concerns, a horizontal safety distance $d_S=2$ m between any individual and the UAV is needed. To take this into account and at the same time design a practical (and thus a smooth) inspection path, we leverage our proposed approach that relies on (\ref{eq:safe-dist-barycenter})--(\ref{eq:safe-dist-normal}).
        
        \Figure[t](topskip=0pt, botskip=0pt, midskip=0pt)[width=0.99\linewidth]{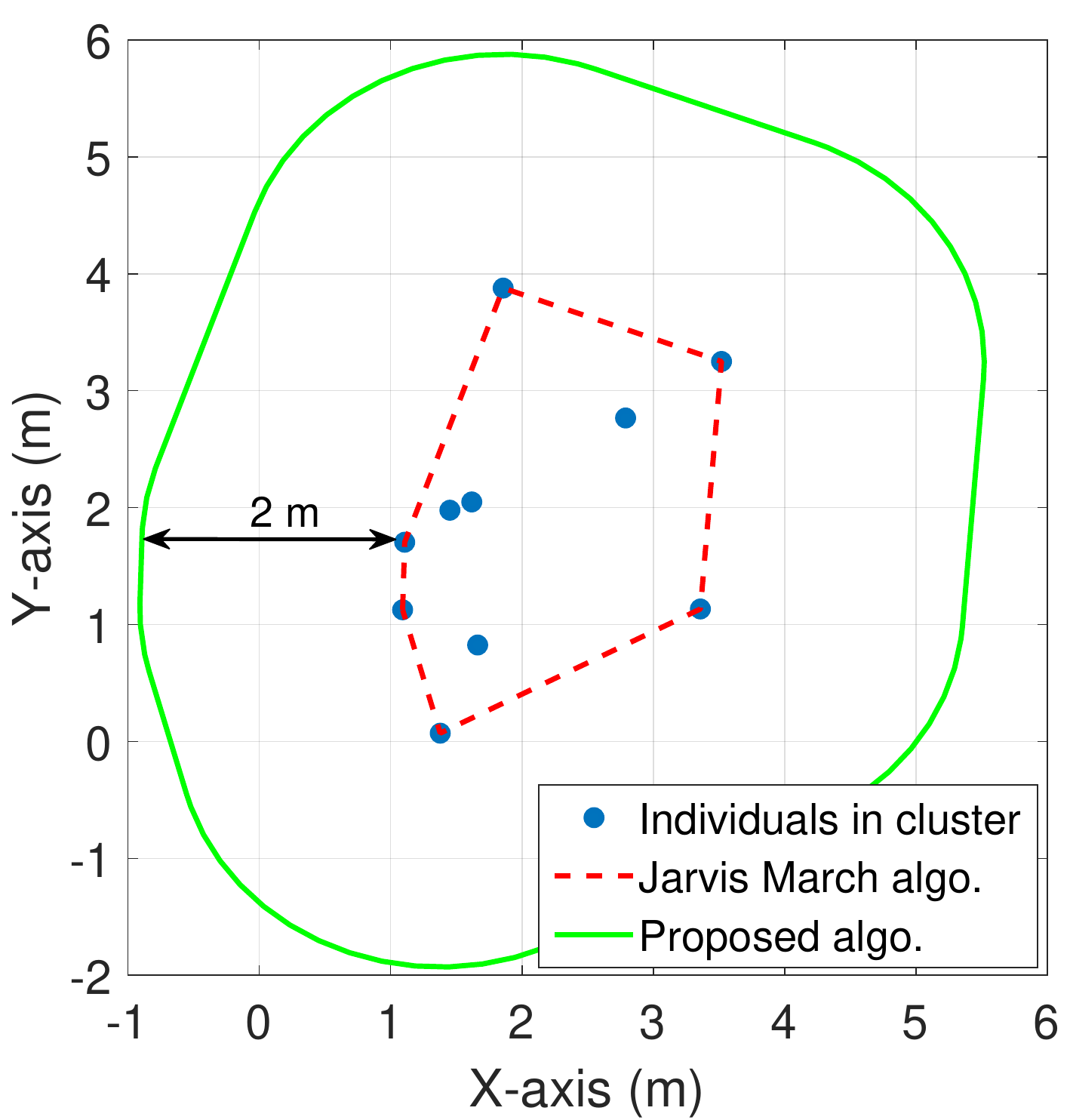}{Cluster inspection path design $(d_s = 2 \; m)$.\label{fig:cvx-hull-sim}}
        
        Finally, the overall UAV trajectory is a combination of the main and cluster inspection paths. In Fig. \ref{fig:uav-full-path}, we present the complete UAV trajectory to inspect clusters of people, obtained through processing with our proposed framework (i.e., human detection, boundig box correction, coordinates mapping, clustering, and UAV trajectory design) image 0000011\_00234\_d\_0000001 of the VisDrone2019 dataset, shown here as Fig. \ref{fig:visdrone-samples}a. In Fig. \ref{fig:uav-full-path}, we mention the infection risk score of each cluster. Finally, given $\alpha=1$, the UAV trajectory is designed to prioritize clusters with high risk scores.
        
        \Figure[t](topskip=0pt, botskip=0pt, midskip=0pt)[width=0.99\linewidth]{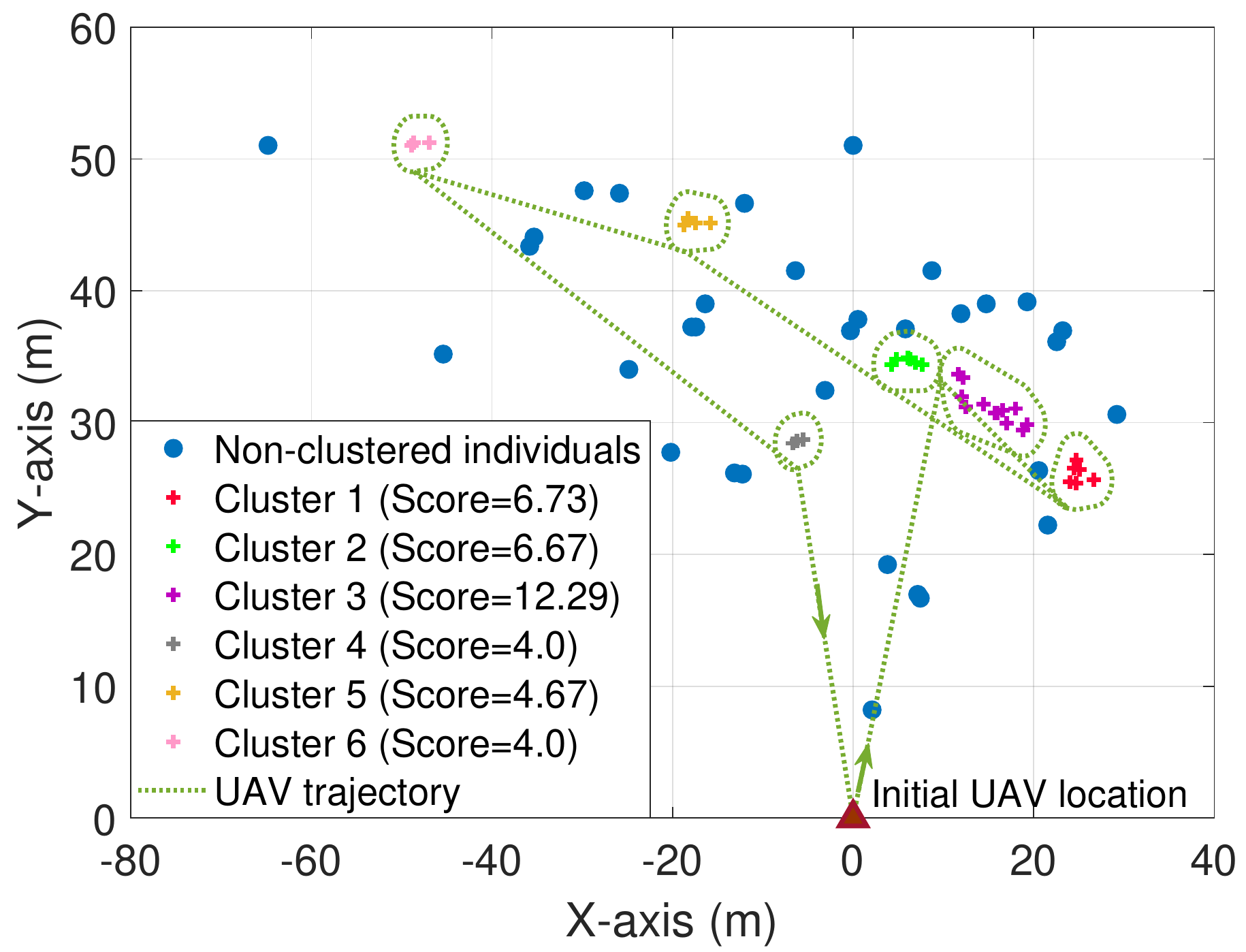}{Output of the crowd monitoring framework (image 0000011\_00234\_d\_0000001 of VisDrone2019 dataset). \label{fig:uav-full-path}}

\section{Conclusion} \label{sec:conc-future-wrks}
    In this paper, we investigated UAV-based crowd monitoring for post COVID-19 outdoor activities. Based on captured images from a flying UAV, we proposed a complete surveillance framework composed of three steps: 1) Human detection and localization in captured images, 2) bounding box correction, coordinates mapping and individuals clustering in a real-world coordinates system, and 3) UAV trajectory planning for further inspection of clusters.  
    The first step was realized using the scaled YOLOv4 approach. The second relied on optical assumptions, the pinhole camera model, and DBSCAN clustering. Finally, the third step was realized based on a number of heuristic and meta-heuristic approaches as well as a novel proposed algorithm for cluster inspection. The obtained results draw the following insights:
    1) Efficient human detection depends on the angle from which the image was captured, 2) coordinates mapping is very sensitive to the estimation error in individuals' bounding boxes drawing, and 3) 2-Opt presented the best performances, in terms of cost and execution time, compared to baseline approaches, and thus is preferred for practical real-time deployments.
    
\bibliographystyle{unsrt}  
\bibliography{tau}

\EOD

\end{document}